\documentclass[lettersize,journal]{IEEEtran}
\usepackage{amsmath,amsfonts}
\usepackage{algorithmic}
\usepackage{algorithm}
\usepackage{array}
\usepackage[caption=false,font=normalsize,labelfont=sf,textfont=sf]{subfig}
\usepackage{textcomp}
\usepackage{stfloats}
\usepackage{url}
\usepackage{verbatim}
\usepackage{graphicx}
\usepackage{cite}
\usepackage{float}
\usepackage{color}
\usepackage{soul}
\usepackage{multirow}
\usepackage{makecell}
\usepackage{bbding}
\usepackage{hyperref}
\usepackage{textcomp}
\usepackage{booktabs} 
\usepackage{mathrsfs}

\begin{document}

\title{V$^2$-SfMLearner: Learning Monocular Depth and Ego-motion for Multimodal Wireless Capsule Endoscopy}

\author{Long Bai, Beilei Cui, Liangyu Wang, Yanheng Li, Shilong Yao, Sishen Yuan, Yanan Wu, Yang Zhang, Max Q.-H. Meng,~\IEEEmembership{Fellow,~IEEE}, Zhen Li, Weiping Ding,~\IEEEmembership{Senior Member,~IEEE}, Hongliang Ren,~\IEEEmembership{Senior Member,~IEEE}
\thanks{The work was supported by Hong Kong RGC CRF C4026-21GF, GRF 14203323, GRF 14216022, GRF 14211420, NSFC/RGC Joint Research Scheme N\_CUHK420/22; Shenzhen-Hong Kong-Macau Technology Research Programme (Type C) Grant 202108233000303; Guangdong Basic and Applied Basic Research Foundation \#2021B1515120035; Shenzhen Key Laboratory of Robotics Perception and Intelligence (ZDSYS20200810171800001), at the Southern University of Science and Technology. 
(Corresponding to: H. Ren.)}
\thanks{$^\dagger$L. Bai, B. Cui, and L. Wang are co-first authors.}
\thanks{L. Bai, B. Cui, L. Wang, S. Yuan, Y. Wu, \& H. Ren are with Dept. of Electronic Engineering, The Chinese University of Hong Kong, Hong Kong, China. (Email: b.long@ieee.org, hlren@ieee.org)}
\thanks{Y. Li \& S. Yao are with City University of Hong Kong, Hong Kong, China.}
\thanks{S. Yao \& M. Meng are with Shenzhen Key Laboratory of Robotics Perception and Intelligence, and Dept. of Electronic and Electrical Engineering, Southern University of Science and Technology, Shenzhen, China.}
\thanks{Y. Zhang is with the School of Mechanical Engineering, Hubei University of Technology, Wuhan, China.}
\thanks{Z. Li is with the Dept. of Gastroenterology, Qilu Hospital of Shandong University, Jinan, China.}
\thanks{W. Ding is with the School of Artificial Intelligence and Computer Science, Nantong University, Nantong, China.}
}

\markboth{Journal of \LaTeX\ Class Files,~Vol.~14, No.~8, August~2021}%
{Shell \MakeLowercase{\textit{et al.}}: A Sample Article Using IEEEtran.cls for IEEE Journals}

\maketitle

\begin{abstract}
Deep learning can predict depth maps and capsule ego-motion from capsule endoscopy videos, aiding in 3D scene reconstruction and lesion localization. However, the collisions of the capsule endoscopies within the gastrointestinal tract cause vibration perturbations in the training data. Existing solutions focus solely on vision-based processing, neglecting other auxiliary signals like vibrations that could reduce noise and improve performance. Therefore, we propose V$^2$-SfMLearner, a multimodal approach integrating vibration signals into vision-based depth and capsule motion estimation for monocular capsule endoscopy. We construct a multimodal capsule endoscopy dataset containing vibration and visual signals, and our artificial intelligence solution develops an unsupervised method using vision-vibration signals, effectively eliminating vibration perturbations through multimodal learning. Specifically, we carefully design a vibration network branch and a Fourier fusion module, to detect and mitigate vibration noises. The fusion framework is compatible with popular vision-only algorithms. Extensive validation on the multimodal dataset demonstrates superior performance and robustness against vision-only algorithms. Without the need for large external equipment, our V$^2$-SfMLearner has the potential for integration into clinical capsule robots, providing real-time and dependable digestive examination tools. The findings show promise for practical implementation in clinical settings, enhancing the diagnostic capabilities of doctors.
\end{abstract}

\def\abstractname{Note to Practitioners}
\begin{abstract}
This paper is motivated by the problem of estimating the depth and ego-motion information for the wireless capsule endoscopy in the human gastrointestinal tract to realize accurate, efficient, robust, and real-time inspection. Our estimation method does not engage any external localization equipment. Instead, inspired by the existing research on integrating capsule endoscopy and inertial measurement units, we introduce vibration signals into vision-based depth and ego-motion estimation approaches, improving the accuracy and robustness of the estimation results based on multimodal learning methods. Research on capsule robots or computer vision can readily be combined with our framework for various clinical and industrial applications.
\end{abstract}


\section{Introduction}

\label{sec:1}

\IEEEPARstart{E}{ach} year, over 28 million patients suffer from gastrointestinal (GI) cancers~\cite{Center2011Global}. It is the second most deadly cancer worldwide~\cite{Arnold2020Global}. Furthermore, cancers in the GI tract tend to vary in symptoms. Diagnosing multiple diseases in the GI tract is quite challenging, because the symptoms among patients are usually different. As a result, physicians cannot use blood tests and symptoms alone to determine the condition and decide on the next treatment steps. The most direct and effective screening method for GI cancers is endoscopy, providing physicians with direct visual information~\cite{bai2024endouic}. Wireless capsule endoscopy (WCE) is a particular type of endoscopy. It does not have to penetrate deep into the GI environment through an external device but comes with its battery, antenna, and imaging equipment~\cite{xu2022evaluation}. When the patient swallows the WCE, it can be controlled remotely (e.g., driven by an external magnetic field~\cite{song2022motion}) or move with the patient's own metabolism until it is expelled from the body, and collects sample information in the patient's GI environment for further analysis, diagnosis and treatment by the physicians. Compared with computerized tomography scanning and traditional endoscopy, WCE has been reported to have better diagnostic sensitivity, less pain and discomfort, and better tolerance during the treatment~\cite{wang2023rethinking,li2023semi}.
Moreover, the flourishing development of deep learning (DL) techniques has further enhanced doctors' ability and speed in reviewing medical images~\cite{yuan2019densely,wang2024surgical,luo2024surgplan}. 

\begin{figure}[tbp] 
	\centering
	\includegraphics[width=0.95\linewidth,scale=1.00]{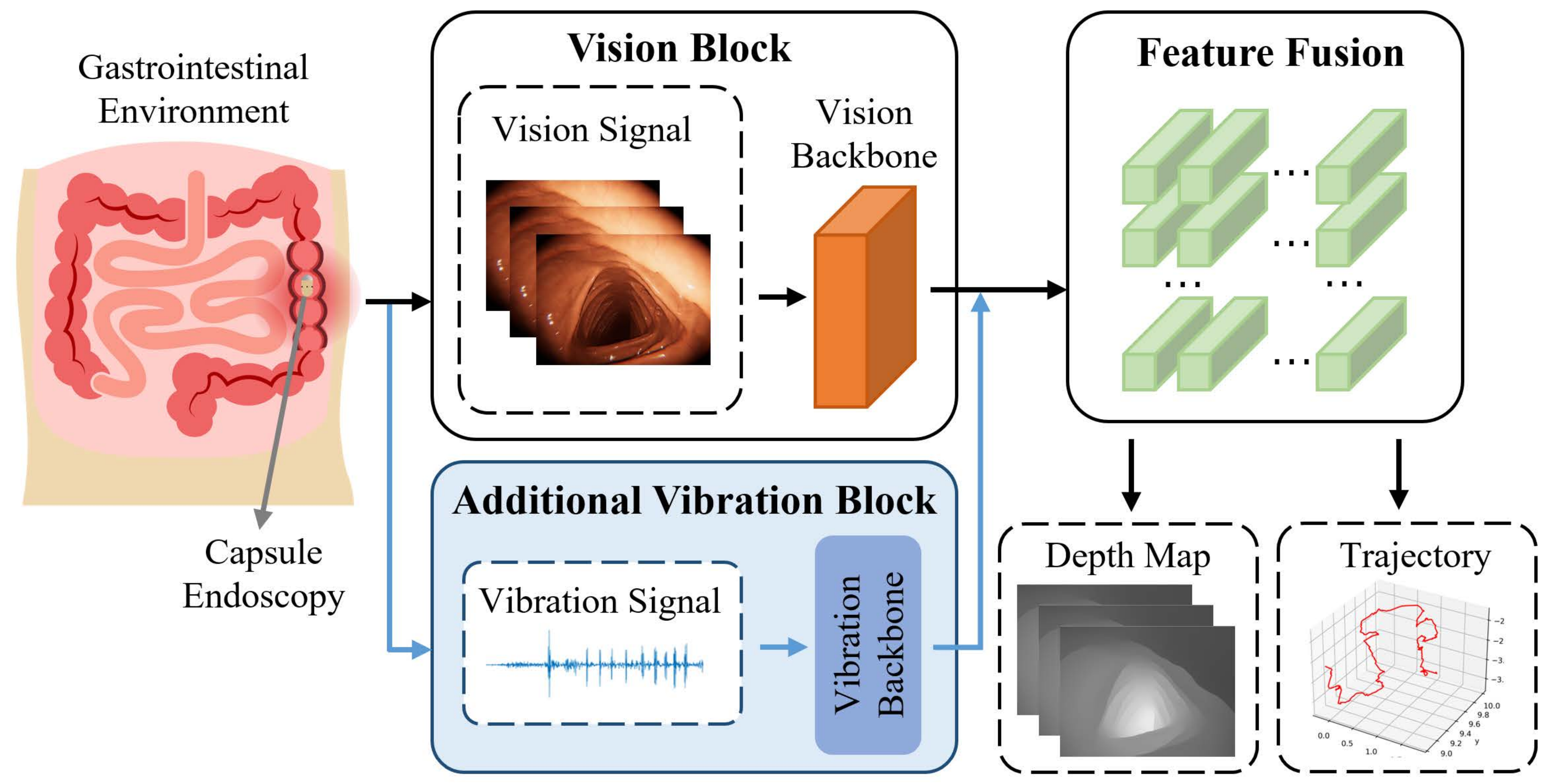}
    \caption{Overview of the vision-vibration framework, against the conventional vision-only depth and ego-motion estimation solution.}
    \label{fig:1}
\end{figure}

Once we have collected enough visual information and identified the presence of a lesion, the subsequent problem is to locate the lesion in the human body~\cite{yuan2016wce,bai2023llcaps}. Magnetic localization methodologies have also been proposed to estimate the real-time WCE ego-motion, while (i) they require large external hardware and cause much more investment of resources~\cite{Son20152d, xu2020novel}; (ii) the localization accuracy is related to the working space. However, strict requirements exist regarding the distance between WCE and magnetic sensors~\cite{wang2020novel}; (iii) environmental magnetic field (e.g., geomagnetic field) has considerable interference with the localization accuracy~\cite{song2021magnetic, wang2021multipoint}.
In this case, researchers have paved a long path in the way of visual odometry utilizing the given visual information, which can provide depth and ego-motion estimation for further diagnosis~\cite{Grasa2013Visual, Mahmoud2018live, Turan2018endovo,huang2024endo,huang2024registering}. Meanwhile, because stereo and structured light endoscopes require larger volumes, higher design complexity, and higher prices, current WCEs are mostly monocular in clinics. This motivates researchers to reconstruct depth information and predict camera ego-motion using monocular visual data.

Nowadays, DL has been numerously integrated with visual odometry. Early methods to learn depth information are supervised strategies~\cite{eigen2014depth}, and they achieved excellent performances. Nevertheless, because of the non-Lambertian reflection and severe noise, it is challenging to acquire a large dataset with ground truth (GT) depth maps via WCE~\cite{AF-SfMLearner}. In this case,  unsupervised solutions to estimate depth and camera motion are sequentially proposed~\cite{bian2019scsfmlearner, fang2020towards, Godard2019Digging, spencer2020defeat, zhou2017sfmlearner}. 
Instead of learning from GT labels, the unsupervised techniques leverage video disparity information for supervision. With the supervisory signal calculated using the visual difference between target and synthesized frames, the network shall be capable of predicting depth information and ego-motion simultaneously.
Researchers have adapted algorithms from autonomous driving to medical endoscopy and can process information from continuous video frames~\cite{RA-Depth, Li2020Unsupervised, EndoSfMLearner, AF-SfMLearner, Sharan2020Domain, yang2024self}. With the depth information to reconstruct the 3D organ and the ego-motion information to acquire the real-time 6D pose, a WCE closed-loop control system can be established, and the problem of lesion localization in the human body can thereby be easily solved.

Vision-only depth and ego-motion estimation methods for monocular WCE often face significant challenges. (1) The movements and collisions of the WCE within the GI tract generate vibrations that cause image jitter, distortion, and artifacts, reducing the accuracy of depth estimation and ego-motion prediction. (2) The GI tract has poor lighting conditions, repetitive textures, and indistinguishable features, making it difficult for vision-only methods to achieve robustness and reliability. These challenges highlight the need for auxiliary information to mitigate the adverse effects of vibration noise and the limitations inherent in vision-only approaches.
Therefore, we envision that if a simple sensor array for acquiring vibration signals can be designed, it will be possible to perform the WCE depth and motion estimation tasks with vibration and visual signals. A simple vibration sensor (e.g., IMU) would not cause hardware redundancy while providing significant information for joint estimation. Furthermore, incorporating vibration signals does not introduce hardware redundancy or complexity, as simple IMUs are lightweight, compact, and already used in robotics and capsule endoscopy research~\cite{abu2015vision, li2022external}. This makes our proposed multimodal framework not only effective but also feasible for clinical integration. The vibration signal can play a vital role in helping to capture the jitter and collision of WCE in the GI tract.

It is observed that the video signal captured by WCE tends to carry some vibration noise~\cite{Liu2013Design}. Vibration signals naturally encode information about the jitter, collision, and movement of the capsule in the GI tract. By capturing these signals through simple inertial sensors (e.g., IMUs), the vibration signals allow us to directly detect and quantify the perturbations caused by capsule vibration, which can then be used to correct the visual data and enhance depth and motion estimation accuracy. Therefore, by integrating vibration signals with visual data, we can effectively denoise the visual features and improve robustness against the challenges mentioned above. We can consider the vibration noise a special type of adversarial signal that is caused by real-world WCE scenarios. Vibration is regarded as an attack on the original image data~\cite{kumar2020adversarial}, and we may defend the attack based on the noise detection mechanism~\cite{gupta2021determining}.

As shown in Fig.~\ref{fig:1}, we propose a solution to fuse visual and vibration signals for learning depth maps and camera ego-motion. The solution consists of a vision branch, a vibration branch, a Fourier heterogeneous fusion module, and prediction decoders for depth maps and camera ego-motion, respectively. The vibration branch and fusion module are plug-and-play components, designed to be easily integrated into existing vision-based algorithms.
We conduct experiments in a virtual GI environment named VR-Caps~\cite{Incetan2021Vrcaps}. VR-Caps provides realistic WCE simulations, and its synthetic data has been shown to generalize well to real-world scenarios, helping address data and label shortages~\cite{Incetan2021Vrcaps, EndoSfMLearner, hofer2021sim2real}. The main contributions of our work are summarized as follows:

\begin{itemize}
    \item We propose V$^2$-SfMLearner, the first unsupervised vision-vibration framework for monocular WCE to predict depth and ego-motion. The framework contains a specific vibration network branch and a Fourier heterogeneous fusion module, designed to be compatible with vision-only algorithms.
    \item We design a novel Fourier heterogeneous (FH) fusion module to combine visual and vibration features in the frequency domain. This module uses SNR estimation from vibration signals to suppress noise in visual features, while the MLSTM-based vibration network processes high-dimensional vibration data to extract meaningful features. Together, they address challenges like vibration noise and low-texture GI environments in WCE.
    \item Our experiments demonstrate that the proposed framework outperforms SOTA vision-only algorithms, achieving superior performance and robustness in depth map reconstruction and ego-motion prediction. Additionally, we plan to publicly release our dataset, which includes depth and ego-motion ground truth, and the vibration signal data, to support future research in this domain.
    \item Our vibration network branch and FH module are cost-effective and clinically feasible. By leveraging readily available vibration sensors, our solution avoids the need for external hardware like magnetic sensors, reduces system complexity, and can be easily integrated into current WCE devices for real-world clinical applications.
\end{itemize}

\section{Related Work}
\label{sec:2}

Researchers first used traditional multiview stereo algorithms to generate the 3D GI scene and estimate the camera ego-motion, e.g., shape-from-motion (SfM)~\cite{Leonard2018Evaluation}, and simultaneous localization and mapping (SLAM)~\cite{sun2020plane}. SfM has been applied on sinus surgeries with sparse bundle adjustment~\cite{Leonard2018Evaluation}. Besides, Grasa \emph{et al.}~\cite{Grasa2013Visual} utilized SLAM to generate the 3D abdominal cavity reconstruction on monocular laparoscope image sequences. Mahmoud \emph{et al.}~\cite{Mahmoud2018live} reported a fast, robust, and dense SLAM method with frame clusters that provide a larger parallax of parallax from the motion of the endoscopy.
However, the light source in the endoscopy environment is usually insufficient, and the GI environment always changes with the body. The endoscopy images also lack sufficient distinguishable features. The above all make it increasingly difficult to extract features manually; in this case, plenty of mismatches will happen when conducting feature matching. Therefore, the performance of the existing multiview stereo algorithms is far from perfect. 

Recently, DL-based algorithms have attempted to infer depth and ego-motion information directly from large-scale WCE data. Turan \emph{et al.}~\cite{Turan2018endovo} designed Deep EndoVO, a monocular WCE visual odometry using recurrent networks. Mahmood \emph{et al.}~\cite{Mahmood2018Deep} targeted the poor contrast of lesion topography in the colon. They presented a supervised mono-endoscopy depth prediction with DL and conditional random field. 
However, as mentioned in Section~\ref{sec:1}, since obtaining an abundance of labeled data in real WCE is quite challenging, supervised learning can hardly get practical applications. Sharan \emph{et al.}~\cite{Sharan2020Domain} attempted to adapt a self-supervised solution in autonomous driving - Monodepth~\cite{Godard2019Digging}, for stereoscopic endoscopic depth estimation for mitral valve surgery. Ozyoruk \emph{et al.}~\cite{EndoSfMLearner} propounded unsupervised Endo-SfMLearner, combining residual networks and spatial attention to focus on distinguishable texture tissue features. Li \emph{et al.}~\cite{Li2020Unsupervised} further integrated the temporal features between consecutive frames to boost the performance. Moreover, Yang \emph{et al.}~\cite{yang2024self} integrated convolutional neural networks and transformers for endoscopic depth estimation, while the works of He \emph{et al.}~\cite{he2024monolot} and Zhang \emph{et al.}~\cite{zhang2023lite} discussed the development of lightweight methods for monocular endoscopic depth estimation. Cui \emph{et al.} constructed fully supervised~\cite{cui2024surgical} and self-supervised~\cite{cui2024endodac} depth estimation paradigms for endoscopy by fine-tuning natural foundation models~\cite{oquab2023dinov2,yang2024depth}.

Although current DL-based vision algorithms can overcome the difficulties of traditional methods in feature extraction, the existing algorithms still need to fully consider and utilize all available information. Recent works have revealed the feasibility and necessity of integrating multiple data modalities to further enhance the performance of estimation tasks with DL~\cite{xu2022information}. Nevertheless, based on our current search and reading, rarely researcher has ever considered introducing vibration signals to predict depth and ego-motion estimation for capsule robots. Abu~\emph{et al.}~\cite{abu2015vision} has preliminarily validated the feasibility of fusing visual and inertial information for reconstructing 3D scenes in capsule endoscopy. However, they still lack a complete pipeline for 3D reconstruction and pose estimation, and their work has not been validated on large-scale datasets. Our fusion framework is expected to defend the vibration perturbations, establish a comprehensive prediction pipeline, and further improve the estimation performance.

\section{Methodology}
\label{sec:3}

\begin{figure*}[t]
    \centering
    \includegraphics[width=0.96\textwidth, trim=0 230 135 0]{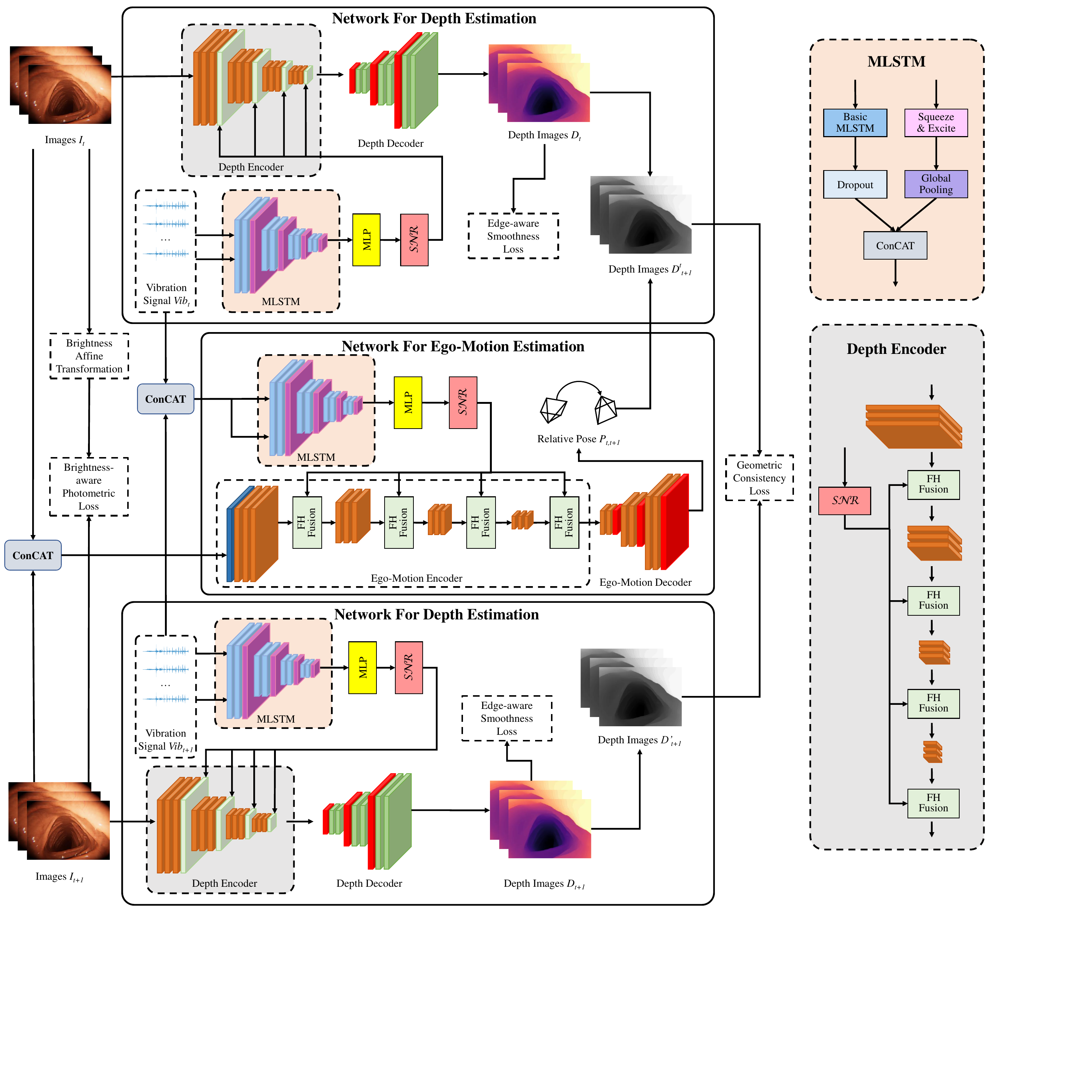}
    \caption{
    The network architecture of the V$^2$-SfMLearner framework. 
    Continuous unlabeled images ($I_t$, $I_{t+1}$) and vibration signals (${Vib}_t$, ${Vib}_{t+1}$) are respectively sent to the network for depth estimation, and predict dense disparity maps ($D_t$, $D_{i+1}$). Meanwhile, the network for ego-motion estimation shall predict the relative WCE pose $P_{i,i+1}$. The output of the vibration branch will feed the Fourier heterogeneous (FH) fusion module after each vision encoder block. The predicted depth map is warped based on the WCE ego-motion information to obtain $D^i_{i+1}$. The pixel-wise disparity between $D^i_{i+1}$ and the interpolated depth map $D'_{i+1}$ is calculated by geometry consistency loss.
    The detailed structures of MLSTM and the depth encoder are presented on the right. FH fusion denotes the Fourier heterogeneous fusion module.
    }
    \label{fig:2}
\end{figure*}

\subsection{Learning Depth Maps and WCE Motion} 
\label{sec:3-1}
Self-supervised depth estimation mainly relies on the photometric and geometric constraint between a target frame and a reprojected frame from the source frame to the target frame. Formally, with known camera parameters, the right images $I_r$ are mapped to $I_{r'}$ according to Equ.~(\ref{eq:3-1-1}). The depth estimation neural network $D$ is trained and optimized via Equ.~(\ref{eq:3-1-2}), where the optimization objective is the difference between reprojected image $I_{r'}$ and left image $I_l$.
\begin{equation}
    \left\{
    \begin{aligned}
    & I_{r'} = I_{r} \left(D(\rho^r) + \rho^r \right) \\
    & I_{l'} = I_{l} \left(D(\rho^l) \right)
    \end{aligned} \right. ,
\label{eq:3-1-1}
\end{equation}

\begin{equation}
    \theta^* = \mathop{\arg\min}\limits_{\theta} \sum_j\sum_i \mathcal{L} \left(I_j^{r'} (\rho_i^r;\theta), I_j^{l'} (\rho_i^l;\theta)\right),
    \label{eq:3-1-2}
\end{equation}
Here, $\rho_i^r$ and $\rho_i^l$ represent the $i$th pixel of the $j$th right view $I_j^r$ and left view $I_j^l$, respectively.
$\theta$ is the network parameter in the depth estimation neural network $D(\cdot)$. $\mathcal{L}$ denotes the loss function, which will be described in Equ.~(\ref{eq:3-1-6})-(\ref{eq:3-1-11}).
Subsequently, based on the obtained optimal neural network $D^*(\cdot)$, the depth $d_i$ of pixels $\rho_i$ in the image $I$ can be predicted with $d_i = (f \cdot B)/D(p_i)$. 
$i$ represents $i$th pixel $\rho_i$ in the image $I$. 
$f$ represents the focal length, and $B$ is the inter-camera distance.
$f$ and $B$ can be calculated with known camera parameters. 

Nevertheless, for video-based monocular estimation, image pairs are not available. In this case, two continuous video frame images $I_t$ and $I_{t+1}$ at $t$ moment and $t+1$ moment will serve as the image pairs to feed the deep neural network. However, Equ.~(\ref{eq:3-1-1}) does not hold for monocular depth estimation now because the camera pose will change with time $t$. 
Godard~\emph{et al.}~\cite{godard2017unsupervised} introduced an ego-motion estimation model $P(\cdot)$ predicting camera ego-motion: inter-frame spatial displacement $(x,y,z)$ and angular movement $(\varphi, \vartheta, \psi)$.
This method projects the previous frame's depth map to the next frame using ego-motion info. The required image pair becomes the projected previous frame and the current frame.
Thus, the inter-frame depth map minimization is optimized through the ego-motion network, resolving inconsistent camera poses between frames.

Thus, Equ.~(\ref{eq:3-1-1}) could be modified to Equ.~(\ref{eq:3-1-4}), with the two networks $D(\cdot)$ and $P(\cdot)$ optimized by Equ.~(\ref{eq:3-1-5}):
\begin{equation}
    \left\{
    \begin{aligned}
    & I_{t'} = I_{t} \left(KP_{[t \to t+1]}(\rho_t) D(\rho_t) K^{-1}\rho_t \right) \\
    & I_{t+1'} = I_{t+1} \left(D(\rho_{t+1}) \right)
    \end{aligned} \right. ,
    \label{eq:3-1-4}
\end{equation}
\begin{small}
\begin{equation}
\begin{aligned}
    & \theta_D^*, \theta_P^* = \mathop{\arg\min}\limits_{\theta_D,\theta_P} \sum_j\sum_i \mathcal{L} \left(I_j^{t'} (\rho_i^t;\theta_D,\theta_P), I_j^{t+1'} (\rho_i^{t+1};\theta_D)\right),
    \label{eq:3-1-5}
\end{aligned}
\end{equation}
\end{small}
Here $P_{[t \to t+1]}(\rho_t)$ denotes the camera motion at the $t$ moment. $K$ denotes the camera's internal reference matrix. $\theta^*$ denotes the model parameters at convergence. 
Collaborative training for depth and ego-motion is essential to obtain accurate prediction results.

\subsection{Learning Objective} 
We construct our loss function based on the weighted combination of three components: brightness awareness photometric loss $\mathcal{L}_p$~\cite{EndoSfMLearner}, edge-aware smoothness loss $\mathcal{L}_s$~\cite{godard2017unsupervised}, and geometric consistency loss $\mathcal{L}_g$~\cite{bian2019scsfmlearner}. 
Firstly, the brightness-aware photometric loss $\mathcal{L}_p$ consists of $\mathcal{SSIM}$ loss, $\mathcal{L}_2$-norm, and brightness consistency loss:
\begin{equation}
\begin{aligned}
    \mathcal{L}_p  = \frac{\epsilon}{2} & \left(1-\mathcal{SSIM}(\mathbf{T}_{[t\to t+1]}(I_{t'}),I_{t+1'})\right) \\
    & + (1-\epsilon) ||\mathbf{T}_{[t\to t+1]}(I_{t'})-I_{t+1'}||_2,
    \label{eq:3-1-6}
\end{aligned}
\end{equation}
where the weight $\epsilon$ is set to 0.85. The brightness consistency loss~\cite{EndoSfMLearner} is combined to reduce the adverse effect of the brightness difference between continuous image frames. 
$\mathbf{T}_{[t\to t+1]}(\cdot)$ denotes the brightness transformation, which transforms the brightness of $I_t$ to match the brightness of $I_{t+1}$ at time $t$.

\begin{equation}
    \mathbf{T}_{[t\to t+1]}(I_t) = a_{[t\to t+1]} I_t + c_{[t\to t+1]},
    \label{eq:3-1-8}
\end{equation}
where $a_{[t\to t+1]}$ and $c_{[t\to t+1]}$ are the brightness affine transformation parameters.

Next, we employ the edge-aware smoothness loss $\mathcal{L}_s$ to ensure the edge region prediction rationality:
\begin{equation}
    \mathcal{L}_s = |\partial_x d_t^*|e^{-|\partial_x I_t|} + |\partial_y d_t^*|e^{-|\partial_y I_t|},
    \label{eq:3-1-7}
\end{equation}
$d_t^*={d_t}/{\bar{d_t}}$ represents the normalized average inverse depth to avoid depth reduction. Here, $x$ and $y$ denote the horizontal and vertical pixel coordinates, respectively. $\partial_x$ and $\partial_y$ denote pixel differences in $x$ and $y$ directions.

Finally, the geometric consistency loss $\mathcal{L}_g$ is employed to minimize the depth difference of the projected images:
\begin{equation}
    \mathcal{L}_g = \sum_i 
    \frac{|D_{t+1}^{t}(\rho_i)-D_{t+1}'(\rho_i)|}{D_{t+1}^{t}(\rho_i)+D_{t+1}'(\rho_i)},
    \label{eq:3-1-9}
\end{equation}

Here, $D_{t+1}^{t}(\rho)$ represents the depth of $I_{t+1}$ predicted and mapped from $I_t$, while $D_{t+1}'(\rho)$ denotes the depth map from $I_{t+1}$. Consequently, the ultimate loss function shall be:
\begin{equation}
    \mathcal{L} = \alpha \mathcal{L}_p + \beta \mathcal{L}_s + \gamma \mathcal{L}_g ,
\label{eq:3-1-11}
\end{equation}
where $\alpha$, $\beta$ and $\gamma$ are weights for each loss.

\subsection{Network Structure}
\label{sec:3-2}

Both depth estimation and ego-motion estimation networks employ encoder-decoder structure as in baselines. The input of the whole framework consists of vision and vibration signals. In each training iteration, the vision part contains 3-frame images collected by the camera in the endoscopy. The vibration part is the 6-channel 240-point vibration signal collected by vibration sensors around the WCE. The vibration signal corresponds to 40 temporal sampling points around the first frame of 3-frames.

\subsubsection{Vision Network Branch}
Our vision encoder is similar to the baselines. The only difference is that the fusion mapping module (details in Section~\ref{sec:3-3}) is added after each encoder block to blend the vibration signal. 
As an illustration, consider the baseline model EndoSfMLearner~\cite{EndoSfMLearner}. Within this model, the encoder is comprised of a Max-pooling layer and multiple residual blocks~\cite{he2016deep}. Each residual block has a residual connection, a convolution layer, batch norm, and ReLU.

\subsubsection{Vibration Network Branch}
We employ the Multiplicative Long Short-Term Memory (MLSTM)~\cite{mlstm} model as our vibration network branch. The vibration branch includes a basic LSTM sub-branch and an attention sub-branch containing the squeeze-and-excite blocks~\cite{hu2018squeeze}. 
Finally,  the outputs from both sub-branches are amalgamated through the utilization of a concatenation operation, and the vibration branch will return a vector containing high-dimensional information about the vibration signal.

\subsubsection{Prediction Decoders}
The prediction decoders are designed for depth estimation and ego-motion estimation, respectively. We follow the decoders in each vision baseline algorithm to ensure that our vibration branch and fusion modules have strong compatibility with existing vision algorithms. For example, in the baseline model EndoSfMLearner~\cite{EndoSfMLearner}, the decoder for depth estimation contains five decoder blocks and a final decision module, respectively, and its output is the corresponding depth map. Each decoder block contains two convolution layers and two ELU activation functions. The decoder for ego-motion estimation is a single feed-forward network with an output of a 6-dimension vector representing the 6-Dof pose. The 6-Dof pose is then transformed into a $4 \times 4$ transformation matrix for reprojection illustrated in Equ.~(\ref{eq:3-1-4}).

\subsection{Fusion strategy} 
\label{sec:3-3}

\begin{figure}[t]
    \centering
    \includegraphics[width=0.9\linewidth, trim=-80 800 920 0]{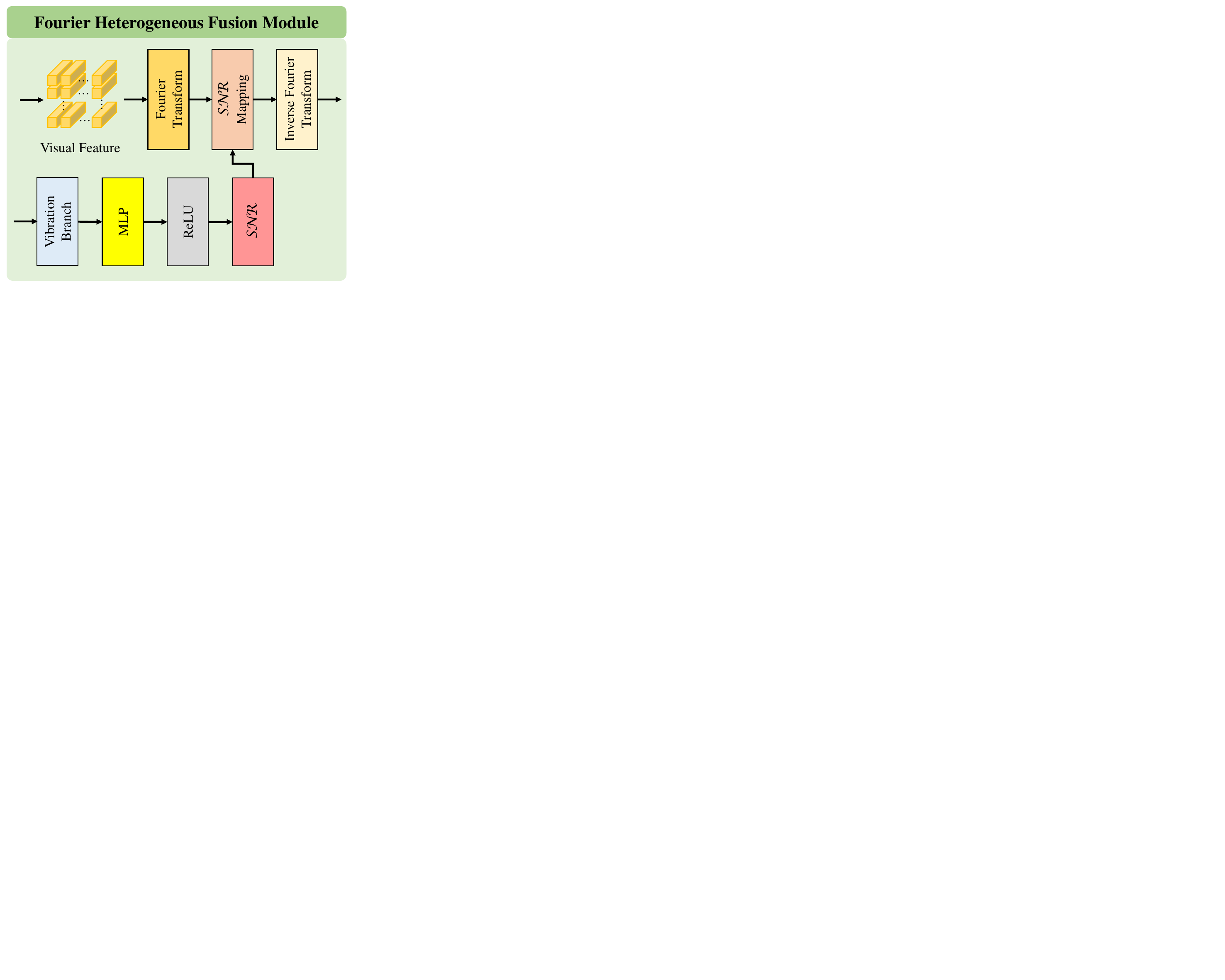}
    \caption{Fourier heterogeneous fusion module. The $\mathcal{SNR}$ of the vibration signal is obtained by the MLP after the vibration branch. The visual feature map is output from each vision encoder block and is then Fourier transformed. The $\mathcal{SNR}$ is fed into the visual feature map in the Fourier domain to remove vibration noise. Subsequently, the visual feature map shall feed the following vision encoder block or decoder after the inverse Fourier domain transform.}
    \label{fig:fusion}
\end{figure}

Our fusion strategy allows for the convenient insertion of the vibration signal network into the vision signal network. 
The structure of our fusion module is depicted in the lower right corner of Fig.~\ref{fig:2}.
The main challenge of this module is that as one-dimensional timing information, the vibration signal does not have spatial information, so it is difficult to directly combine the vibration signal feature and RGB image feature into the network. 
To tackle this challenge, we specifically develop a Fourier heterogeneous fusion module, to modulate the corresponding vision and vibration signals from different modalities. 
The vision signal is transformed into the Fourier domain, and the joint heterogeneous representation of the visual signal and the vibration signal is subsequently established.

Firstly, assuming that the vibration interference received by the image is additive noise, the signal with noise can be expressed as:
\begin{equation}
    F_l(x_{vi};\theta_{vi}) = h_l(x_{vi}) * F_l'(x_{vi};\theta_{vi}) + n_l(x_{vi}),  \label{equ:1}
\end{equation}
where the $F'(x_{vi};\theta_{vi})$ represents the ideal pure vision signal without noise, $F(x_{vi};\theta_{vi})$ represents the original information with noise, in our case, the output information after each encoder. $h(x_{vi})$ represents information channel, $n(x_{vi})$ represents noise, which is not related to vision signal $F'(x_{vi};\theta_{vi})$. $l$ denotes the $l$-th encoder. Thus, we can explore the best mutual representation of visual and vibration signals. A Fourier transform of the above equation yields:
\begin{equation}
    F_l(\xi) = H_l(\xi) * F_l'(\xi) + N_l(\xi),  \label{equ:2}
\end{equation}
\begin{equation}
    F_l^*(\xi) = H_l^*(\xi) * {F_l'}^*(\xi) + N_l^*(\xi),  \label{equ:3}
\end{equation}
where $F(\xi)$, $F'(\xi)$, $H(\xi)$, $N(\xi)$ represent the frequency domain representation of $F(x_{vi};\theta_{vi})$, $F'(x_{vi};\theta_{vi})$, $h(x_{vi})$, $n(x_{vi})$, and the $F^*(\xi)$, $F'^*(\xi)$, $H^*(\xi)$, $N^*(\xi)$ represent the conjugate of $F(\xi)$, $F'(\xi)$, $H(\xi)$, $N(\xi)$, respectively.

Therefore we can obtain:
\begin{equation}
    F_l'(x_{vi};\theta_{vi}) = \mathscr{F}^{-1} \left\{\frac{F_l(\xi) H_l^*(\xi)}{H_l(\xi)H_l^*(\xi) + \frac{1}{\mathcal{SNR}}} \right\}, \label{equ:4}
\end{equation}
\begin{equation}
    \mathcal{SNR}_{org} = \frac{F_l(\xi)F_l^*(\xi)}{N_l(\xi)N_l^*(\xi)},
\end{equation}
where the notation $\mathscr{F}^{-1} {\cdot}$ signifies the inverse Fourier transform, and $\mathcal{SNR}_{org}$ represents the signal-to-noise ratio for current vision signal. 
Therefore, the pure vision signal can be jointly represented by the original input vision signal and $\mathcal{SNR}_{org}$, and we can then easily recover the pure vision signal without noise based on the $\mathcal{SNR}_{org}$ for current vision signal. However, it is also challenging to achieve the $\mathcal{SNR}_{org}$ directly from the images. 
Without the pure vision signal as the GT, the deep neural network can hardly extract the $\mathcal{SNR}_{org}$ only from vision signals. 
In this case, we consider the indirect estimation of the $\mathcal{SNR}_{org}$ of the image from the vibration signal. Specifically, we calculate the $\mathcal{SNR}$ based on the vibration signal, then modulate the vision signal in Fourier space with the $\mathcal{SNR}$. 
As outlined in Section~\ref{sec:3-2}, we obtain the output feature vector from the vibration signal encoder. 
Then, we define certain  $\mathcal{SNR}$ grades during training and use an MLP layer to map it into $\mathcal{SNR}$: 
\begin{equation}
    \mathcal{SNR} = ReLU\{W_l^T G_l(x_{vib};\theta_{vib}) + b_l\}, \label{equ:5}
\end{equation}
where $G_l(x_{vib};\theta_{vib})$ is the output feature vector of the vibration signal encoder, $\mathcal{SNR}$ is the output of the MLP, the terms $W$ and $b$ correspond to the weight and bias employed by the MLP layer, and $ReLU$ is the nonlinear activation function of the MLP layer. 
As our entire framework operates in an unsupervised manner, we do not impose any supervision constraints on the $\mathcal{SNR}$. Instead, we optimize the $\mathcal{SNR}$ together with the entire network architecture through Equ.~(\ref{eq:3-1-11}).
Thus, a modulation mapping module is added after each vision signal encoder block to map the predicted noise with the output features, and therefore, to pure the feature maps without noise. 

The entire process of the module can be seen in \textbf{Algorithm}~1. 
The inputs to the Fourier heterogeneous fusion module consist of two components: the output feature map originating from the vibration signal encoder, and the output feature map produced by each vision signal encoder.
Therefore, the fused and pure feature map is obtained via the FH fusion module, thus achieving the effect of driving the overall network update. The obtained feature map will further feed the decoder mentioned in Section~\ref{sec:3-3} to conduct depth and ego-motion estimation tasks. We have thus effectively established a defense mechanism against vibration perturbations.

\begin{algorithm}[t] 
\label{alg:1}
\caption{Fourier Heterogeneous Fusion Module} 
    {\bf Input:}
    (1) output feature map of each vision signal encoder $F_l(x_{vi};\theta_{vi})$,
    (2) output feature map of vibration signal encoder $G_l(x_{vib};\theta_{vib})$ \\
    {\bf Output:}
    Feature map $F_l'(x_{vi};\theta_{vi})$ \\
    1: \hspace*{0.1in} {\bf Initialization}:  $ \theta_{vi}, \theta_{vib} \gets 0$, $\{W$\}, $\{b$\} \\
    2: \hspace*{0.1in} {\bf Repeat} not converged \\
    3: \hspace*{0.1in} \hspace*{0.1in} {\bf For} the $l^{th}$ vision signal encoder block \\
    4: \hspace*{0.1in} \hspace*{0.1in} \hspace*{0.1in} Estimate SNR using output feature map of vibration signal encoder: \\ \hspace*{0.25in} \hspace*{0.1in} \hspace*{0.1in} $\mathcal{SNR} \gets ReLU\left\{W_l^T G_l(x_{vib};\theta_{vib}) + b_l\right\}$ \\
    5: \hspace*{0.1in} \hspace*{0.1in} \hspace*{0.1in} Calculate $F_l(\xi)$ using Fourier transform: \\ 
    \hspace*{0.25in} \hspace*{0.1in} \hspace*{0.1in} $F_l(\xi) \gets \mathscr{F} \left\{F_l(x_{vi};\theta_{vi}) \right\}$ \\
    6: \hspace*{0.1in} \hspace*{0.1in} \hspace*{0.1in} Calculate $H(\xi)$ using Fourier transform: \\ 
    \hspace*{0.25in} \hspace*{0.1in} \hspace*{0.1in} $H(\xi) \gets \mathscr{F} \left\{W_l\right\}$ \\
    7: \hspace*{0.1in} \hspace*{0.1in} \hspace*{0.1in} Calculate the conjugate of $H(\xi)$: \\ 
    \hspace*{0.25in} \hspace*{0.1in} \hspace*{0.1in} $H^*(\xi) \stackrel{conjugate}{\longleftarrow} H(\xi)$ \\
    8: \hspace*{0.1in} \hspace*{0.1in} \hspace*{0.1in} Calculate $F_l'(\xi)$ using Equ. (\ref{equ:4}): \\ 
    \hspace*{0.25in} \hspace*{0.1in} \hspace*{0.1in} $F_l'(\xi) \gets \frac{F_l(\xi) H_l^*(\xi)}{H_l(\xi)H_l^*(\xi) + \frac{1}{\mathcal{SNR}}} $ \\
    9: \hspace*{0.1in} \hspace*{0.1in} \hspace*{0.1in} Calculate $F_l'(x_{vi};\theta_{vi})$ using Fourier inversion: \\ 
    \hspace*{0.25in} \hspace*{0.1in} \hspace*{0.1in} $F_l'(x_{vi};\theta_{vi}) \gets \mathscr{F}^{-1} \left\{F_l'(\xi) \right\}$ \\
    10: \hspace*{0.05in} \hspace*{0.1in} {\bf End For}
\end{algorithm}

\subsection{Feasibility \& Compatibility of Vision-Vibration Fusion}

\subsubsection{Hardware Feasibility}
The integration of vibration sensors/IMUs into capsule robots for hardware and spatial dimension feasibility is underscored by advancements in sensor fusion techniques and miniaturization. Abu-Khei \emph{et al.}~\cite{abu2015vision} outlined a method using visual and inertial data fusion to map the GI tract, emphasizing the feasibility and robustness of incorporating an IMU system within the capsule's limited space. Li~\emph{et al.}~\cite{li2022external} introduced a novel localization approach combining external magnetic field sensing with internal inertial sensing for 6-DOF pose estimation of a magnetic WCE, demonstrating the feasibility of embedding sophisticated sensors without requiring complex structures or specific motions. These studies collectively illustrate the technological progress in miniaturizing and efficiently integrating sensors into capsule robots, enabling enhanced diagnostic capabilities and navigation precision within constrained spatial dimensions.

\subsubsection{Compatibility with Vision-only Methods}
Our vision-vibration framework is highly compatible with most existing vision-only methods because the proposed Fourier Heterogeneous Fusion Module can be seamlessly integrated into many networks. To be specific, a vibration feature map is obtained with the MLSTM module and the size should be aligned with the vision feature map generated by the network's encoder. Then, the proposed Fourier Heterogeneous Fusion Module can be applied based on the estimated SNR from the vibration signal, which does not change the size of the vision feature maps. Once we obtain the vibration signals, our proposed method can be used as a plug-and-play component that is easily embeddable in any depth prediction architecture with a small number of extra trainable parameters.

\subsubsection{The Use of Virtual Environment}
Our dataset is generated in a simulated environment due to the challenges of collecting ground truth depth, ego-motion, and vibration signals in real GI scenes with current clinical technologies. Simulation enables us to provide accurate annotations for depth and motion, which are essential for training and evaluation. Specifically, we utilize VR-Caps, a high-fidelity simulation platform that replicates the GI environment with realistic textures, lighting, and dynamics, including vibration noise and peristalsis. VR-Caps has been validated in prior studies~\cite{ruano2024leveraging, liu2023sparse, pore2022colonoscopy, zhang2022deep, rodriguez2022tracking, ahmad20233d}, demonstrating strong generalization of models trained on its data to real-world scenarios. For example, Pore \textit{et al.}~\cite{pore2022colonoscopy} and Ahmad \textit{et al.}~\cite{ahmad20233d} involved professional doctors to confirm its realism, while Ruano \textit{et al.}~\cite{ruano2024leveraging} showed comparable depth estimation performance in real endoscopic settings. Although simulated data was used, we validated the robustness of our unsupervised multimodal algorithm through extensive experiments. These results demonstrate the reliability of our method and its readiness for deployment in real-world environments.

\section{Experiments}
\label{sec:4}

\subsection{Dataset}
\label{sec:4-1}

Since the existing WCE datasets~\cite{EndoSfMLearner} do not provide the vibration signal, and obtaining the depth and ego-motion GT from real GI scenes is quite difficult, we produce a new dataset via VR-Caps\footnote{\href{https://github.com/CapsuleEndoscope/VirtualCapsuleEndoscopy}{github.com/CapsuleEndoscope/VirtualCapsuleEndoscopy}}~\cite{Incetan2021Vrcaps} which is a virtual capsule endoscopy environment with advanced rendering techniques. The VR-Caps setup provides a variety of digestive tract organ models. Multiple research works have proven datasets collected from VR-Caps hold excellent generalization ability, and can achieve accuracy comparable to training on real datasets when tested on real scenarios~\cite{Incetan2021Vrcaps, EndoSfMLearner}. Besides, some vision-based depth and ego-motion estimation work also validate their proposed methodologies in virtual datasets~\cite{Li2020Unsupervised}. 
Our dataset is collected in the above simulation environment with one monocular WCE. To collect the vibration signals, we further establish the vibration sensors in the VR-Caps environment, and bound the vibration sensors to the monocular WCE. Therefore, six strings of one-dimensional vibration signals can be collected (following the form of an IMU).

\begin{figure}[t]
	\centering
	\includegraphics[width=0.8\linewidth, scale=1, trim=0 10 75 0]{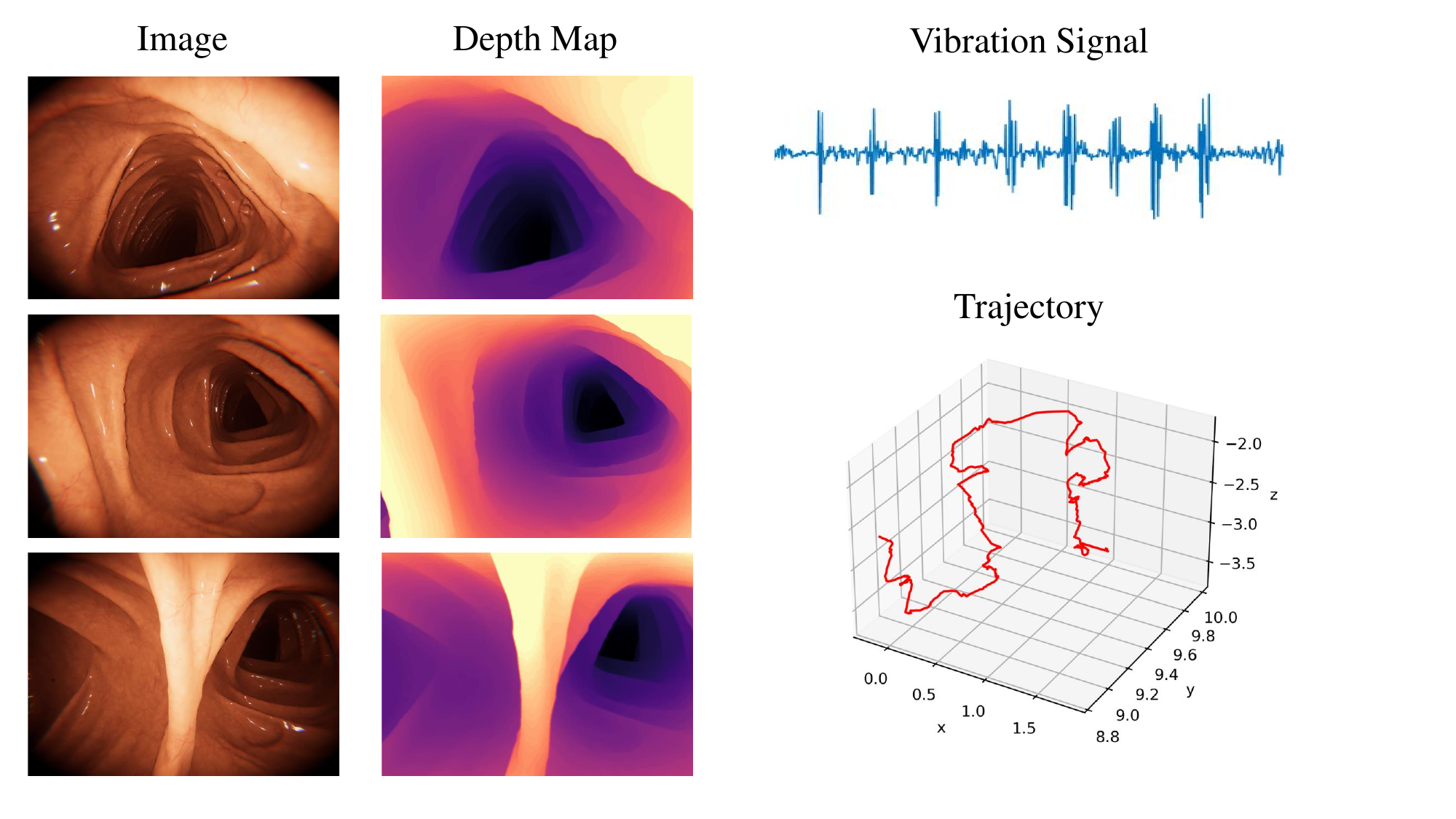}
    \caption{Overview of our datasets. Left: original images; Middle: the depth GT; Right Top: vibration signal example; Right Bottom: camera ego-motion example.
    }
    \label{fig:4}
\end{figure}

Vibration noise may originate from the natural movements of the GI tract, electronic interference, and the mechanical operations of the capsule itself. Gaussian noise is a mathematical model that effectively represents the statistical properties of many physical phenomena, including vibration. Real-world vibration noise in capsule endoscopy can be characterized by random fluctuations that, over time, approximate the statistical profile of Gaussian noise.
In this case, to simulate the vibration of an endoscope, we add various intensities of Gaussian noise to the movement of the capsule in three spatial dimensions following~\cite{wu2017vins}. The noise intensity is within the level $1$ to level $5$, and the Gaussian noise is within $[-1, 1]$. The example of the collected dataset is shown in Fig.~\ref{fig:4}. 
The image data is collected at 3 FPS by following~\cite{hong2012comparison}, while the vibration signals were sampled at a frequency of 40 Hz after carefully reviewing existing real-world setups~\cite{li2022external,zhang2022multimagnetometer}. The vibration signals are normalized before feeding the vibration network branch.
We invite two WCE experts to manually control the capsule in VR-Caps setup, and name the datasets collected by different experts as Multimodal-WCE-1 (MM-WCE-1) and MM-WCE-2. We divide the ``Train" dataset with 6 video sequences into training and validation sets as $8:2$. The test set contains 5 separate video sequences. 
All datasets are collected with the vibration intensity label, GT depth maps, and WCE ego-motion.

\subsection{Implementation Details}

\label{sec:4-2}
We integrate our V$^2$-SfMLearner framework with four depth and ego-motion estimation benchmarks: EndoSfMLearner~\cite{EndoSfMLearner}, AF-SfMLearner~\cite{EndoSfMLearner}, RA-Depth~\cite{RA-Depth}, and EndoDAC~\cite{cui2024endodac}. 
The baselines and our fused models are trained by Adam with $400$ epochs. The batch size is $4$, and the learning rate is $1 \times 10^{-4}$. We utilize NVIDIA A100 with Python PyTorch. 
During training, 3 frames of images with 40 vibration signal points constitute the training data at time $t$. Then, the model is tested frame by frame during evaluation.
The three weights of Equ.~(\ref{eq:3-1-11}) are assigned as: $\alpha=1$, $\beta=0.1$, and $\gamma=0.5$.

\begin{table}[h]
    \renewcommand\arraystretch{0.4}
    \small
	\caption{Metrics for ego-motion prediction.
	}
	\centering
	\label{tab:motionmetric} 
        \resizebox{0.45\textwidth}{!}{	
	\setlength{\tabcolsep}{5mm}{
	\begin{tabular}{cc}
		\noalign{\smallskip}\hline\noalign{\smallskip}
		Metric & Definition \\
            \noalign{\smallskip}\hline\noalign{\smallskip}
            ATE & $\frac{\sum_{j=1}^{\mathrm{f}-1} \sqrt{\left(\hat{x}_j-x_j\right)^2+\left(\hat{y}_j-y_j\right)^2+\left(\hat{z}_j-z_j\right)^2}}{\mathrm{f}-1}$\\
		\noalign{\smallskip}\hline\noalign{\smallskip}
            AbsDiff$_t$ & $\frac{1}{\mathrm{f}-1}\sum_{j=1}^{\mathrm{f}-1} \left(|\hat{x_j} - x_i|+|\hat{y_j} - y_i|+|\hat{z_j} - z_i|
            \right)$\\
            \noalign{\smallskip}\hline\noalign{\smallskip}
            AbsRel$_t$ &  $\frac{1}{\mathrm{f}-1}\sum_{j=1}^{\mathrm{f}-1} \left(\frac{|\hat{x_j} - x_i|}{x_i}+\frac{|\hat{y_j} - y_i|}{y_i}+\frac{|\hat{z_j} - z_i|}{z_i}\right)$\\            
            \noalign{\smallskip}\hline\noalign{\smallskip}
            ARE &  $\frac{\sum_{j=1}^{\mathrm{f}-1} \sqrt{\left(\hat{\varphi}_j-\varphi_j\right)^2+\left(\hat{\vartheta}_j-\vartheta_j\right)^2+\left(\hat{\scriptsize{\psi}}_j-\scriptsize{\psi}_j\right)^2}}{\mathrm{f}-1} $ \\
            \noalign{\smallskip}\hline\noalign{\smallskip}
            AbsDiff$_r$ & $\frac{1}{\mathrm{f}-1}\sum_{j=1}^{\mathrm{f}-1} \left(|\hat{\varphi_j} - \varphi_i|+|\hat{\vartheta_j} - \vartheta_i|+|\hat{\psi_j} - \psi_i|\right)$\\
            \noalign{\smallskip}\hline\noalign{\smallskip}
            AbsRel$_r$ & $\frac{1}{\mathrm{f}-1}\sum_{j=1}^{\mathrm{f}-1} \left(\frac{|\hat{\varphi_j} - \varphi_i|}{\varphi_i}+\frac{|\hat{\vartheta_j} - \vartheta_i|}{\vartheta_i}+\frac{|\hat{\psi_j} - \psi_i|}{\psi_i}\right)$ \\
            \noalign{\smallskip}\hline
	\end{tabular}}}
\end{table}

\subsection{Evaluation metrics}
\label{sec:4-3}

For depth estimation, we adopt five metrics from~\cite{zhou2017sfmlearner} as follows: the absolute relative error (AbsRel), the square relative error (SqRel), the root-mean-squared error (RMSE), the root-mean-square logarithmic error (logRMSE), and Accuracy (Acc) with a threshold as $1.25$. Furthermore, we use the absolute difference (AbsDiff$=(1/N) \sum_i^N |D_i - \hat{D_i}|$) to assess the absolute error between GT and predicted depth maps. $N$ represents the pixel number. $D_i$ and $\hat{D_i}$ represent the GT and predicted depth value at the $i^{th}$ pixel, respectively.

\begin{figure*}[t]
    \centering
	\includegraphics[width=0.9\linewidth,scale=1, trim=0 20 0 0]{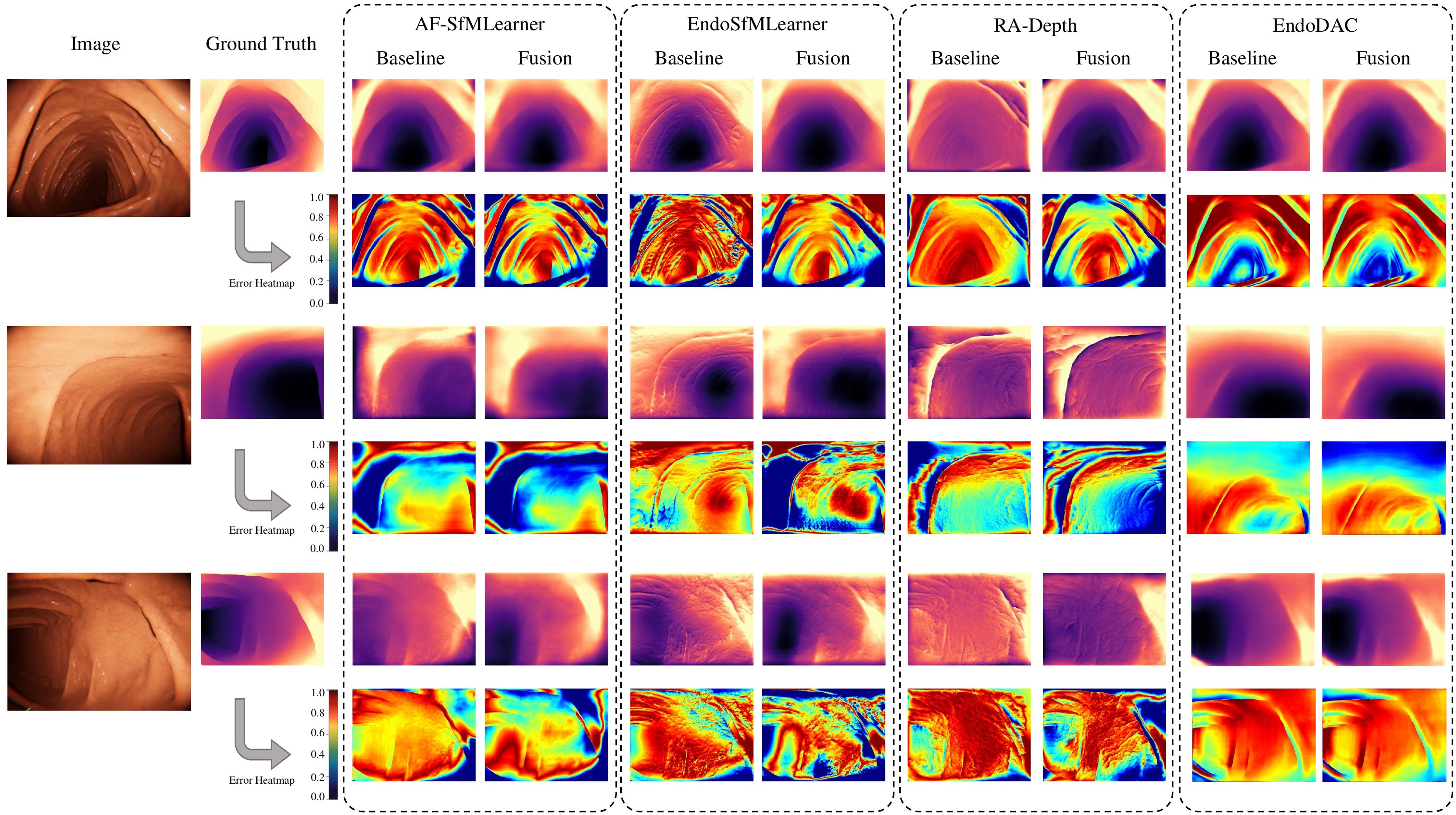}
    \caption{The qualitative experimental results of depth estimation of our fusion framework, against vision-only baselines EndoSfMLearner~\cite{EndoSfMLearner}, AF-SfMLearner~\cite{AF-SfMLearner}, RA-Depth~\cite{RA-Depth}, and EndoDAC~\cite{cui2024endodac}. The error heat maps are calculated with the normalized difference between the GT and the predicted depth map. Blue represents low error, and red represents high error.}
    \label{fig:depth}
\end{figure*}

For ego-motion estimation, we employ the absolute trajectory errors, including the absolute translational error (ATE) and the absolute rotational error (ARE). Besides, we extend ATE into the absolute translational difference (AbsDiff$_t$) and the absolute relative translational error (AbsRel$_t$), and extend ARE into the absolute rotational difference (AbsDiff$_r$) and the absolute relative rotational error (AbsRel$_r$). Our ego-motion metrics are defined in Table~\ref{tab:motionmetric},
in which $\mathrm{f}$ denotes the frames of a WCE video, and $j$ represents the sequence number containing two frames from the WCE video. $x, y, z$ represent the spatial position in three dimensions, and $\varphi, \vartheta, \psi$ represent Euler angles. $\hat{x}_j, \hat{y}_j, \hat{z}_j$ are the predicted position difference between two adjacent frames, and $\hat{\varphi}_j, \hat{\vartheta}_j, \hat{\psi}_j$ are the predicted angle difference between two adjacent frames. $x_j, y_j, z_j, \varphi_j, \vartheta_j, \psi_j$ are the corresponding GT.

\begin{table*}[h]
    \renewcommand\arraystretch{0.4}
    \small
	\caption{Depth and ego-motion estimation performance comparison of our vision-vibration model against vision-only baselines EndoSfMLearner~\cite{EndoSfMLearner}, AF-SfMLearner~\cite{AF-SfMLearner}, RA-Depth~\cite{RA-Depth}, and EndoDAC~\cite{cui2024endodac} on the MM-WCE-1 dataset. The column "\textbf{Vibration}" indicates whether the network contains the vibration branch.
	}
	\centering
	\label{tab:depth_wce1}  
    \resizebox{\textwidth}{!}{	
	\begin{tabular}{c|c|cccccc|cccccc}
		\noalign{\smallskip}\bottomrule[1pt]\noalign{\smallskip}
		\multirow{2}{*}{Methods} & \multirow{2}{*}{Vibration} & \multicolumn{6}{c|}{Depth Estimation} & \multicolumn{6}{c}{Ego-motion Estimation} \\
        \noalign{\smallskip}\cline{3-14}\noalign{\smallskip}
        & & AbsDiff $\downarrow$ & AbsRel $\downarrow$ & SqRel $\downarrow$ & RMSE $\downarrow$ & logRMSE $\downarrow$ & Acc $\uparrow$ & ATE $\downarrow$ & AbsDiff $_t \downarrow$ & AbsRel$_t  \downarrow$ & ARE $\downarrow$ &AbsDiff $_r \downarrow$ & AbsRel$_r  \downarrow$  \\
		\noalign{\smallskip}\hline\noalign{\smallskip}
	
        \multirow{2}{*}{\makecell[c]{EndoSfMLearner~\cite{EndoSfMLearner}}}
		& \scriptsize{\color{red}\XSolidBrush} & 0.2862 & 0.1125 & 0.0635 & 0.3931 & \textbf{0.1481} & 86.87 & \textbf{0.1097} & \textbf{0.0480} & 0.1348 & 0.2157 & 0.0668 & 0.5196 \\\noalign{\smallskip}
		& \scriptsize{\color{green}\CheckmarkBold} & \textbf{0.2690} & \textbf{0.1065} & \textbf{0.0578} & \textbf{0.3925} & 0.1515 & \textbf{88.53} & 0.1232 & 0.0515 & \textbf{0.1236} & \textbf{0.1677} &  \textbf{0.0582} & \textbf{0.4812}
          \\

		\noalign{\smallskip}\hline\noalign{\smallskip}

        \multirow{2}{*}{\makecell[c]{AF-SfMLearner~\cite{AF-SfMLearner}}}
		& \scriptsize{\color{red}\XSolidBrush} & 0.3197 & 0.1248 & 0.0769 & 0.4494 & 0.1725 & 83.72 & 0.4857 & 0.3166 & 0.2370 & 0.1363 & 0.0555 & 0.4905 \\ \noalign{\smallskip}
		& \scriptsize{\color{green}\CheckmarkBold} & \textbf{0.2811} & \textbf{0.1081} & \textbf{0.0617} & \textbf{0.3999} & \textbf{0.1514} & \textbf{87.47} & \textbf{0.1606} & \textbf{0.0880} & \textbf{0.1450} & \textbf{0.1269} & \textbf{0.0521} & \textbf{0.4592} \\
  
		\noalign{\smallskip}\hline\noalign{\smallskip}

        \multirow{2}{*}{\makecell[c]{RA-Depth~\cite{RA-Depth}}}
		& \scriptsize{\color{red}\XSolidBrush} & 0.3423 & 0.1418 & 0.0897 & 0.4555 & 0.1802 & 79.50 & 2.3928 & 1.2515 & 0.5341 & 0.2713 & \textbf{0.0889} & 0.7226\\\noalign{\smallskip}
		& \scriptsize{\color{green}\CheckmarkBold} & \textbf{0.3310} & \textbf{0.1347} & \textbf{0.0860} & \textbf{0.4498} & \textbf{0.1763} & \textbf{81.56} & \textbf{2.0251} & \textbf{1.0439} & \textbf{0.5355} & \textbf{0.2039} & 0.0900 & \textbf{0.7189} \\

        \noalign{\smallskip}\hline\noalign{\smallskip}
        
        \multirow{2}{*}{\makecell[c]{EndoDAC~\cite{cui2024endodac}}}
		& \scriptsize{\color{red}\XSolidBrush} &  0.2836 & 0.1170 & 0.0706 & 0.3949 & 0.1522 & 87.78 & 0.1067 & 0.0512 & 0.0687 & 0.1342 & 0.0576 & 0.9291\\\noalign{\smallskip}
		& \scriptsize{\color{green}\CheckmarkBold} & \textbf{0.1838}  & \textbf{0.0728}  & \textbf{0.0312}  & \textbf{0.2711}  & \textbf{0.1047}  & \textbf{95.04} & \textbf{0.0678} & \textbf{0.0328} & \textbf{0.0662} & \textbf{0.1281} & \textbf{0.0550} & \textbf{0.9064} \\
            \bottomrule[1pt]
	\end{tabular}}
\end{table*}

\subsection{Results}
\label{sec:3-4}

The results of our multimodel solution are analyzed quantitatively and qualitatively based on four SOTA vision-only depth and ego-motion estimation methods: EndoSfMLearner~\cite{EndoSfMLearner}, AF-SfMLearner~\cite{AF-SfMLearner}, RA-Depth~\cite{RA-Depth}, and EndoDAC~\cite{cui2024endodac}. We followed the original setups in the vision network branch and prediction decoders, and further combined their methods with our vibration network branch and the proposed Fourier heterogeneous fusion module. 

\begin{table*}[t]
    \renewcommand\arraystretch{0.4}
    \small
	\caption{Depth and ego-motion estimation performance comparison of our vision-vibration model against vision-only baselines EndoSfMLearner~\cite{EndoSfMLearner}, AF-SfMLearner~\cite{AF-SfMLearner}, RA-Depth~\cite{RA-Depth}, and EndoDAC~\cite{cui2024endodac} on the MM-WCE-2 dataset.
	}
	\centering
	\label{tab:depth_wce2}  
    \resizebox{\textwidth}{!}{		
	\begin{tabular}{c|c|cccccc|cccccc}
		\noalign{\smallskip}\bottomrule[1pt]\noalign{\smallskip}
		\multirow{2}{*}{Methods} & \multirow{2}{*}{Vibration} & \multicolumn{6}{c|}{Depth Estimation} & \multicolumn{6}{c}{Ego-motion Estimation} \\
        \noalign{\smallskip}\cline{3-14}\noalign{\smallskip}
        & & AbsDiff $\downarrow$ & AbsRel $\downarrow$ & SqRel $\downarrow$ & RMSE $\downarrow$ & logRMSE $\downarrow$ & Acc $\uparrow$ & ATE $\downarrow$ & AbsDiff $_t \downarrow$ & AbsRel$_t  \downarrow$ & ARE $\downarrow$ &AbsDiff $_r \downarrow$ & AbsRel$_r  \downarrow$  \\
		\noalign{\smallskip}\hline\noalign{\smallskip}
	
        \multirow{2}{*}{\makecell[c]{EndoSfMLearner~\cite{EndoSfMLearner}}}
		& \scriptsize{\color{red}\XSolidBrush} & 0.2792 & 0.1114 & 0.0630 & 0.3831 & \textbf{0.1461} & 0.8553 & 0.0985 & 0.1502 & 0.0103 & 0.1578 & 0.5332 & \textbf{1.1832} 
 \\\noalign{\smallskip}
		& \scriptsize{\color{green}\CheckmarkBold} & \textbf{0.2615} & \textbf{0.1049} & \textbf{0.0575} & \textbf{0.3820} & 0.1488 & \textbf{0.8691} & \textbf{0.0952} & \textbf{0.1394} & \textbf{0.0093} & \textbf{0.1359} & \textbf{0.4990} & 1.1872 \\
		\noalign{\smallskip}\hline\noalign{\smallskip}

        \multirow{2}{*}{\makecell[c]{AF-SfMLearner~\cite{AF-SfMLearner}}}
		& \scriptsize{\color{red}\XSolidBrush} & 0.3210 & 0.1263 & 0.0788 & 0.4490 & 0.1728 & 81.94 & \textbf{0.3676} & 0.2447 & 0.0274 & \textbf{0.1741} & 0.6617 & 1.5334 
 \\\noalign{\smallskip}
		& \scriptsize{\color{green}\CheckmarkBold} & \textbf{0.3167} & \textbf{0.1230} & \textbf{0.0736} & \textbf{0.4370} & \textbf{0.1663} & \textbf{83.17} & 0.4032 & \textbf{0.2273} & \textbf{0.0258} & \textbf{0.1741} & \textbf{0.6472} & \textbf{1.5230} \\
 
		\noalign{\smallskip}\hline\noalign{\smallskip}

        \multirow{2}{*}{\makecell[c]{RA-Depth~\cite{RA-Depth}}}
		& \scriptsize{\color{red}\XSolidBrush} & 0.3244 & 0.1352 & 0.0849 & 0.4364 & 0.1731 & 79.24 & 2.3440 & 0.4903 & 0.4301 & \textbf{0.1820} & 1.4919 & 3.7914 
 \\\noalign{\smallskip}
		& \scriptsize{\color{green}\CheckmarkBold} & \textbf{0.3123} & \textbf{0.1276} & \textbf{0.0800} & \textbf{0.4286} & \textbf{0.1683} & \textbf{81.30} & \textbf{1.8944} & \textbf{0.4851} & \textbf{0.2855} & 0.1836 & \textbf{1.4897} & \textbf{3.7071} \\
  
  	\noalign{\smallskip}\hline\noalign{\smallskip}

        \multirow{2}{*}{\makecell[c]{EndoDAC~\cite{cui2024endodac}}}
		& \scriptsize{\color{red}\XSolidBrush} &  0.2804 & 0.1139 & 0.0625 & 0.3698 & 0.1440 & 87.82 & 0.1251 & 0.0595 & \textbf{0.1474} & 0.1307 & 0.0543 & 2.1700 \\\noalign{\smallskip}
		& \scriptsize{\color{green}\CheckmarkBold} & \textbf{0.1839} & \textbf{0.0725} & \textbf{0.0263} & \textbf{0.2512} & \textbf{0.0981}  & \textbf{95.83} & \textbf{0.1128} & \textbf{0.0540} & 0.1506 & \textbf{0.1272} & \textbf{0.0526} & \textbf{2.1660} \\

            \bottomrule[1pt]
	\end{tabular}}
\end{table*}

\begin{figure}[t]
    \centering
	\includegraphics[width=0.95\linewidth, scale=1, trim=0 0 0 0]{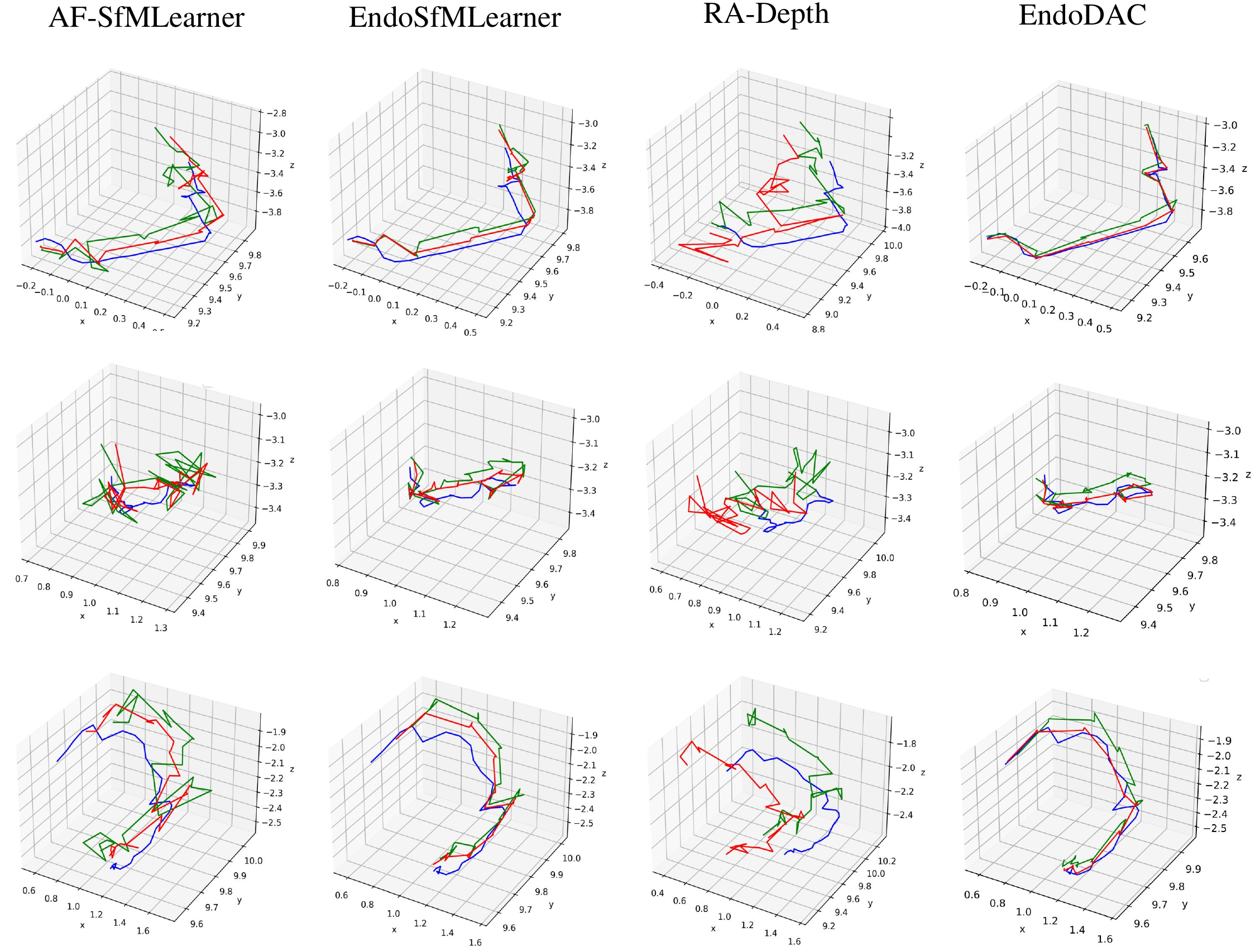}
    \caption{The qualitative experimental results of ego-motion estimation of our fusion framework, against vision-only baselines EndoSfMLearner~\cite{EndoSfMLearner}, AF-SfMLearner~\cite{AF-SfMLearner}, RA-Depth~\cite{RA-Depth}, and EndoDAC~\cite{cui2024endodac}. Blue denotes GT, Green denotes Baseline, and Red denotes our fusion framework.
    }
    \label{fig:motion}
\end{figure}

Firstly, the depth reconstruction performance is presented in Fig.~\ref{fig:depth} qualitatively. The performance of both tasks is shown in Table~\ref{tab:depth_wce1} and~\ref{tab:depth_wce2} quantitatively. 
Our multimodal fusion framework achieves superior performance, realizes less reconstruction error in the depth estimation task, and outperforms all vision-only methods in almost all six evaluation metrics. 
Besides, our solution also qualitatively outperforms the vision-only models, and is more similar to GT depth maps in qualitative comparison.
Specifically, in GT depth maps, the bulge should be shown with a clear separation line from the deeper place, but the depth change from the shallower place should be gradual. However, the texture information displayed by the vision-only EndoSfMLearner~\cite{EndoSfMLearner} in the depth prediction maps is a bright yellow texture line in a dark purple area, which does not match the depth GT. Our method successfully corrects this error and gives depth prediction maps closer to the GT. 
To further prove the superior performance of our framework, we also provide the reconstruction heat maps in Fig.~\ref{fig:depth}, which are calculated with the difference between GT and prediction results. Overall, our fusion method obtains depth maps that have more blue areas on the heat maps, especially evident in the bulge texture in the GI tract. Meanwhile, the depth prediction of the four vision-only methods has specific problems, such as the distortion of Af-SfMLearner~\cite{AF-SfMLearner} around the image edge, the irregular texture of EndoSfMLearner~\cite{EndoSfMLearner}, the poor prediction results of RA-Depth~\cite{RA-Depth} in the middle of the image, and the blurring results of EndoDAC~\cite{cui2024endodac} at the edge of the image. Our fusion framework has improved a series of problems above, and the corresponding areas on the heat maps have also presented more blue proportion, demonstrating the superiority of our vision-vibration framework in depth map reconstruction.

The translational and rotational ego-motion estimations are evaluated, respectively. Our vision-vibration framework still outperforms or performs on par with all SOTA ego-motion estimation methodologies, showing promising ego-motion prediction ability. In qualitative comparisons of trajectories, our method (red) is always the most similar to GT (blue).
Overall, our method is pretty effective in defending against vibration perturbations, and demonstrates superior performance in the unsupervised depth and ego-motion estimation for WCE.

\subsection{Ablation on Vibration Intensity}
\label{sec:3-5}

As we need to estimate the $\mathcal{SNR}$ of the vibration signal for further fusion and prediction, we combine the two datasets as a joint dataset, and conduct an ablation study on different vibration intensities to demonstrate whether our fusion method can improve performance under different intensities of vibration noise. Specifically, we introduce five different levels of random vibration intensities during data collection. For each vibration intensity, we conduct training and validation using data collected under the same condition. We employ Af-SfMLearner~\cite{AF-SfMLearner} as the baseline. The results, as shown in Table~\ref{tab:intensity}, indicate a clear trend: the estimation performance worsens as the vibration intensity increases. However, our fusion framework outperforms vision-only methods in both tasks at each vibration intensity, again demonstrating the superior performance of our fusion framework. These results highlight the significant contribution of the vibration signal in our fusion framework, which plays a critical role in enhancing performance.

\begin{table}[t]
    \renewcommand\arraystretch{0.4}
    \footnotesize
	\caption{
	Ablation study on vibration intensity levels. AF-SfMLearner~\cite{AF-SfMLearner} is employed as baseline in this experiment.
	The column "\textbf{Intensity}" stands for the vibration intensity level.
	}
	\centering
	\label{tab:intensity}  
        \resizebox{0.48\textwidth}{!}{
	\begin{tabular}{c|cc|cc|cc|cc}
        \noalign{\smallskip}\bottomrule[1pt]\noalign{\smallskip}	
        \multirow{2}{*}{AF-SfMLearner} & \multicolumn{4}{c|}{w/o Vibration-Vision Fusion} & \multicolumn{4}{c}{w Vibration-Vision Fusion} \\ \noalign{\smallskip}\cline{2-9}\noalign{\smallskip}     
        & \multicolumn{2}{c|}{Depth} & \multicolumn{2}{c|}{Ego-motion} & \multicolumn{2}{c|}{Depth} & \multicolumn{2}{c}{Ego-motion} \\
        \noalign{\smallskip}\hline\noalign{\smallskip}
        Intensity & AbsDiff $\downarrow$ 
        & Acc $\uparrow$ & ATE $\downarrow$ 
        & ARE $\downarrow$ 
        & AbsDiff $\downarrow$ & Acc $\uparrow$ & ATE $\downarrow$ & ARE $\downarrow$ 
        \\
	\noalign{\smallskip}\hline\noalign{\smallskip}

        $1$ & 0.3109 
        & 0.8403 & 0.3351 
        & 0.1664 
        & \textbf{0.2972}  
        & \textbf{0.8582} & \textbf{0.1873} 
        & \textbf{0.1637} 
        \\\noalign{\smallskip}
        $2$ & 0.3112 
        & 0.8353 & 0.4188 
        & \textbf{0.1698}  
        & \textbf{0.3024} 
        & \textbf{0.8520} & \textbf{0.3147} 
        & 0.1707 
        \\\noalign{\smallskip}
        $3$ & 0.3128 
        & 0.8299 & 0.4535 
        & \textbf{0.1723} 
        & \textbf{0.3088} 
        & \textbf{0.8443} & \textbf{0.3984} 
        & 0.1736 
        \\\noalign{\smallskip}
        $4$ & 0.3165 
        & 0.8236 & \textbf{0.4436} 
        & \textbf{0.1739} 
        & \textbf{0.3120} 
        & \textbf{0.8386} & 0.4684  
        & 0.1750  
        \\\noalign{\smallskip}
        $5$ & 0.3201 
        & 0.8175 & 0.5482 
        & 0.1770  
        & \textbf{0.3179}  
        & \textbf{0.8303} & \textbf{0.4022} 
        & \textbf{0.1766} 
        \\\noalign{\smallskip}

        \bottomrule[1pt]\noalign{\smallskip}
	\end{tabular}}  
\end{table}

\begin{figure*}[!t]
    \centering
    \includegraphics[width=0.9\linewidth, trim=0 0 0 0]{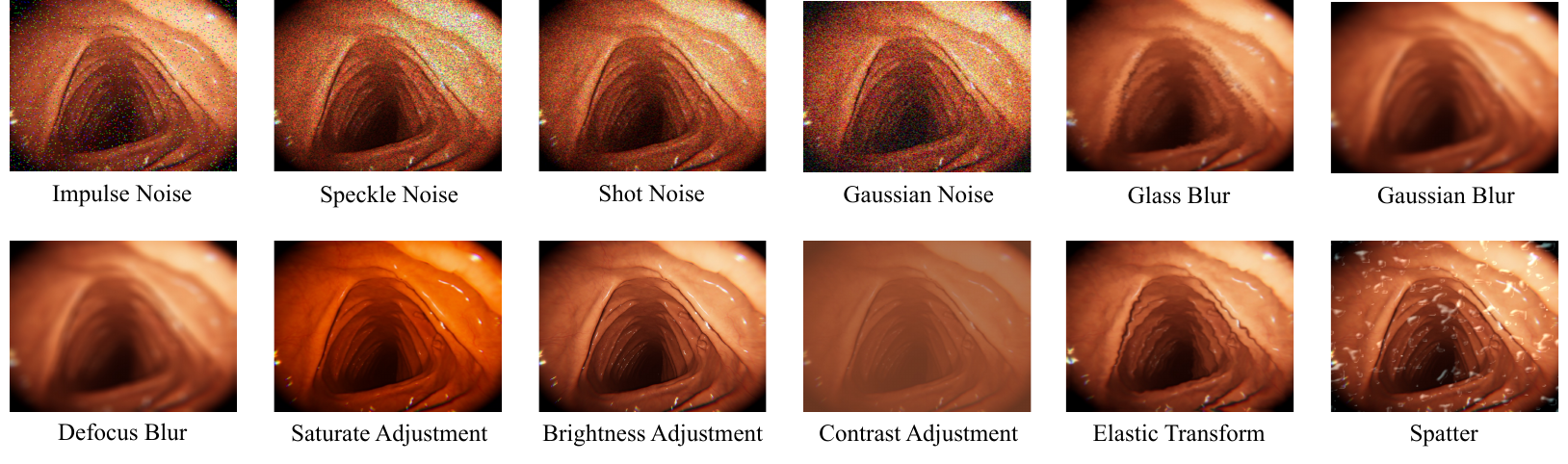}
    \caption{Visualization of the 12 types of corrupted image data for the robustness experiments.}
    \label{fig:robustness_vis}
\end{figure*}

\subsection{Ablation on Fusion Strategy}

As shown in Table~\ref{tab:fusion_method}, we further experiment to investigate whether conventional multimodal fusion methods can effectively integrate vision and vibration signals. Specifically, we replace the denoising fusion strategy with feature-level fusion methods, including concatenation, element-wise summation, gated multimodal fusion~\cite{arevalo2017gmu}, attentional feature fusion (AFF)~\cite{dai2021attentional}, iterative attentional feature fusion (iAFF)~\cite{dai2021attentional}, and bottom-up and top-down attention (BUTD)~\cite{anderson2018bottom}. It is evident that when using these conventional methods to perform various learnable (weighted) combinations of features, the models could not effectively resist interference from vibration noise. Moreover, directly integrating vibration noise with vision data will lead to a decline in feature quality, resulting in an overall decreased performance. As shown in Table~\ref{fig:fusion}, the performances of some conventional feature fusion methods are even lower than the results of vision-only methods, which further demonstrates the effectiveness of our proposed framework.

\subsection{Ablation on Vibration Types}
The vibration may be caused by various reasons therefore we experiment to investigate the performance of our proposed method under different types of vibrations. Two types of vibration, which are the peristalsis of the GI tract and collision with foreign objects (e.g., food residues), are tested as shown in Table~\ref{tab:_vibration_type}. We conduct experiments for different vibrations with the same route and train the model with the same epochs. The performances under Peristalsis vibration are superior in all evaluation metrics for both depth estimation and ego-motion estimation tasks. This may be due to the relatively small amplitude and frequency of gastrointestinal peristalsis, while the direct collision of the capsule with food residues can cause significant movement. Besides, in both scenarios, the models with our proposed vibration-vision fusion strategy obtain better performance also demonstrating the effectiveness of our method against different types of vibration.

\begin{table}[h]
    \renewcommand\arraystretch{0.4}
    \small
	\caption{Experiments on different types of vibration signals. Based on AF-SfMLearner~\cite{AF-SfMLearner}, we compare our proposed framework against vibration caused by peristalsis of the GI tract and collision with foreign objects.}
	\centering
	\label{tab:_vibration_type}  
    \resizebox{0.48\textwidth}{!}{	
	\begin{tabular}{c|cc|cc|cc|cc}
	\noalign{\smallskip}\bottomrule[1pt]\noalign{\smallskip}
        \multirow{2}{*}{AF-SfMLearner} & \multicolumn{4}{c|}{w/o Vibration-Vision Fusion} & \multicolumn{4}{c}{w Vibration-Vision Fusion} \\ \noalign{\smallskip}\cline{2-9}\noalign{\smallskip}     
        & \multicolumn{2}{c|}{Depth} & \multicolumn{2}{c|}{Ego-motion} & \multicolumn{2}{c|}{Depth} & \multicolumn{2}{c}{Ego-motion} \\
        \noalign{\smallskip}\hline\noalign{\smallskip}
        Vibration Type & AbsDiff $\downarrow$ & Acc $\uparrow$ & ATE $\downarrow$ & ARE $\downarrow$ & AbsDiff $\downarrow$ & Acc $\uparrow$ & ATE $\downarrow$ & ARE $\downarrow$\\
	\noalign{\smallskip}\hline\noalign{\smallskip}
  	Peristalsis & 0.3197 & 83.72 & 0.4857 & 0.1363 & \textbf{0.2811}  & \textbf{87.47} & \textbf{0.1606} & \textbf{0.1269}  \\
	\noalign{\smallskip}
  	Collision & 0.3542 & 82.70 & 0.5632 & 0.1431 & \textbf{0.3087} & \textbf{84.98} & \textbf{0.1845} & \textbf{0.1358}  \\

        \bottomrule[1pt]
	\end{tabular}} 
\end{table}

\subsection{Robustness Experiments}
\begin{figure}[!t]
    \centering
    \includegraphics[width=0.9\linewidth, trim=0 120 230 0]{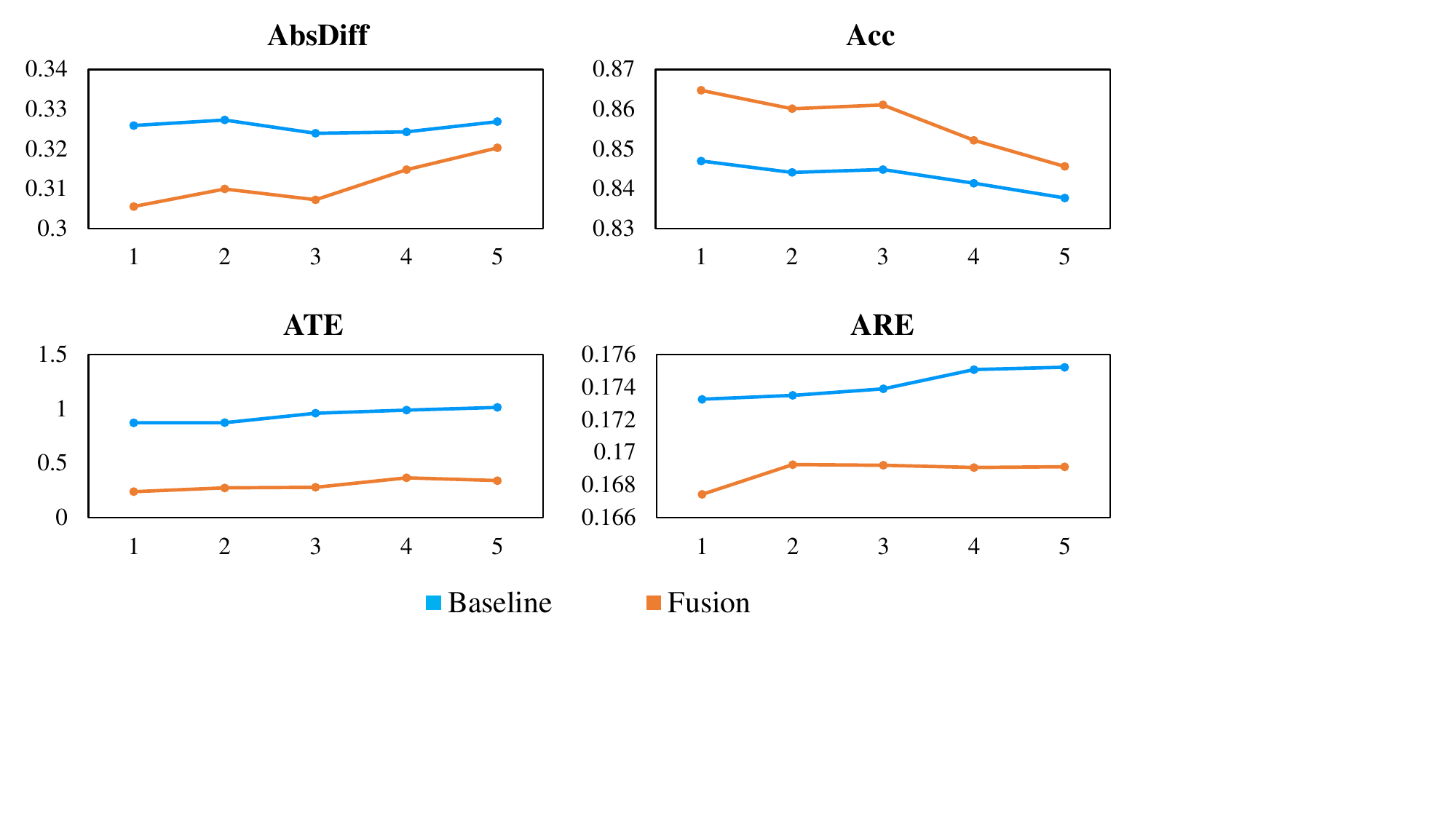}
    \caption{Visualization of the robustness experiment results with different severity levels of our proposed solution against Af-SfMLearner~\cite{AF-SfMLearner}.
    At each severity level, we process the test data with 12 corruption methods and average the prediction results. We present the results with the increasing severity level.
    }
    \label{fig:robustness_level}
\end{figure}

Finally, we conduct a robustness experiment on the joint dataset to observe the model stability when test data is corrupted, and compare the performance between vision-only and vision-vibration framework using Af-SfMLearner~\cite{AF-SfMLearner}. 
We set 12 different types of corruption on the test data based on the severity level from 1 to 5 from the 2D robustness benchmark~\cite{hendrycks2018benchmarking}, as follows:
\begin{itemize}
    \item \textit{Impulse Noise}: Randomly occurring black-and-white pixels, commonly known as ``salt-and-pepper'' noise. In endoscopy, it can simulate pixel-level sensor malfunction or transmission errors caused by the imaging hardware.
    \item \textit{Speckle Noise}: Grainy noise which may simulate texture-like noise on tissue surfaces due to lighting artifacts in endoscopy.
    \item \textit{Shot Noise}: Noise caused by the discrete nature of photons, prevalent in low-light imaging.
    \item \textit{Gaussian Noise}: Fine-grained noise with a normal distribution, often used to model electronic sensor noise or subtle background fluctuations in endoscopic video frames.
    \item \textit{Glass Blur}: A simulated blur effect that mimics the degradation of image quality caused by contamination on the endoscope lens, such as fogging, smearing, or residue.
    \item \textit{Gaussian Blur}: A simulated smoothing effect that mimics image blurring caused by minor camera motion or slight defocusing during endoscopic procedures.
    \item \textit{Defocus Blur}: Blur effect due to the subject being out of focus in the imaging system.
    \item \textit{Saturate Adjustment}: Alteration of color saturation, pushing colors to their extremes or desaturation.
    \item \textit{Brightness Adjustment}: Modification of image brightness, either dimming or brightening the scene.
    \item \textit{Contrast Adjustment}: Changes to the relative difference between dark and light regions in the image.
    \item \textit{Spatter}: Artificial noise resembling splashes or smears, mimicking occlusions or distortions.
    \item \textit{Elastic Transform}: Warping of the image, simulating distortions often seen in real-world settings.
\end{itemize}
Regarding the criteria for selecting these noise types, we aimed to simulate a diverse range of distortions and degradations that are commonly encountered in real-world GI imaging environments. The selected noise types were chosen based on their relevance to challenges posed by: (i) imaging hardware limitations: impulse, Gaussian, and shot noise; (ii) environmental factors: glass, Gaussian, \& Defocus blur, speckle noise; (iii) post-processing effects: saturation, brightness, and contrast adjustments; (iv) physiological conditions and occlusions: spatter, elastic transform. 
As visualized in Figure~\ref{fig:robustness_vis}, the selected noise types align with common challenges faced in GI imaging, such as low-light conditions, defocus issues, and artifacts introduced by camera optics or environmental factors. By covering a broad spectrum of noise and distortion types, we aim to ensure that the framework's performance reflects its potential for practical deployment in clinical and diagnostic settings. The performance on different corruption methods is presented in Table~\ref{tab:robustness_method}, and the performance on each corruption severity level is presented in Fig.~\ref{fig:robustness_level}.

The performance of both vision-only and vision-vibration frameworks decreases with increasing severity levels. However, the vision-vibration-based method shows a significantly slower decline rate and outperforms the vision-only method at every severity level. Meanwhile, in the separate comparison of each corruption method, our method also achieved better performance. Evidently, the vision-only approach suffers from data corruption when estimating the ego-motion, and the ATE is three times that of the vision-vibration framework. Furthermore, our method outperforms all metrics. In summary, two different ways of comparing demonstrate the superior robustness of our vision-vibration fusion method and inspire potential for real-world applications.

\begin{table}[t]
    \renewcommand\arraystretch{0.4}
    \small
	\caption{Ablation study on fusion strategy on the MM-WCE-1 dataset. Based on AF-SfMLearner~\cite{AF-SfMLearner}, we compare our proposed fusion framework against summation, concatenation, Gated Fusion~\cite{arevalo2017gmu}, AFF~\cite{dai2021attentional}, iAFF~\cite{dai2021attentional}, and BUTD~\cite{anderson2018bottom}.}
	\centering
	\label{tab:fusion_method}  
    \resizebox{0.48\textwidth}{!}{	
	\begin{tabular}{c|c|cccccc}
		\noalign{\smallskip}\bottomrule[1pt]\noalign{\smallskip}
            \multirow{2}{*}{Fusion Strategy} & \multirow{2}{*}{Vibration} & \multicolumn{6}{c}{Depth Estimation} \\\noalign{\smallskip}\cline{3-8}\noalign{\smallskip}
            & & AbsDiff $\downarrow$ & AbsRel $\downarrow$ & SqRel $\downarrow$ & RMSE $\downarrow$ & logRMSE $\downarrow$ & Acc $\uparrow$ \\
		\noalign{\smallskip}\hline\noalign{\smallskip}
  	       Concatenation & \scriptsize{\color{green}\CheckmarkBold} & 0.3221 & 0.1308 & 0.0803 & 0.4356 & 0.1702 & 82.38 \\
		\noalign{\smallskip}
  	       Summation & \scriptsize{\color{green}\CheckmarkBold} & 0.3245 & 0.1339 & 0.0846 & 0.4417 & 0.1734 & 81.59 \\
            \noalign{\smallskip} 
	       Gated~\cite{arevalo2017gmu} & \scriptsize{\color{green}\CheckmarkBold} & 0.2914 & 0.1157 & 0.0722 & 0.4140 & 0.1580 & 85.96 \\
		\noalign{\smallskip}        
	       AFF~\cite{dai2021attentional} & \scriptsize{\color{green}\CheckmarkBold} & 0.3196 & 0.1297 & 0.0850 & 0.4403 & 0.1696 & 83.14 \\
		\noalign{\smallskip} 
  	       iAFF~\cite{dai2021attentional} & \scriptsize{\color{green}\CheckmarkBold} & 0.2916& 0.1177 & 0.0716 & 0.4075& 0.1566 & 85.54 \\
		\noalign{\smallskip} 
  	       BUTD~\cite{anderson2018bottom} & \scriptsize{\color{green}\CheckmarkBold} & 0.2875 & 0.1113 & 0.0652 & \textbf{0.3926} & \textbf{0.1482} & 86.42 \\
		\noalign{\smallskip}
  	       Vision-only & \scriptsize{\color{red}\XSolidBrush} & 0.3210 & 0.1263 & 0.0788 & 0.4490 & 0.1728 & 81.94  \\
            \noalign{\smallskip}
		   V$^2$-SfMLearner & \scriptsize{\color{green}\CheckmarkBold} & \textbf{0.2811} & \textbf{0.1081} & \textbf{0.0617} & 0.3999 & 0.1514 & \textbf{87.47}  \\
            \bottomrule[1pt]\noalign{\smallskip}	
            
            \multirow{2}{*}{Fusion Strategy} & \multirow{2}{*}{Vibration} & \multicolumn{6}{c}{Ego-motion Estimation} \\\noalign{\smallskip}\cline{3-8}\noalign{\smallskip}
            & & ATE $\downarrow$ & AbsDiff $_t \downarrow$ & AbsRel$_t  \downarrow$ & ARE $\downarrow$ &AbsDiff $_r \downarrow$ & AbsRel$_r  \downarrow$ \\
		\noalign{\smallskip}\hline\noalign{\smallskip}
  	       Concatenation & \scriptsize{\color{green}\CheckmarkBold} & 0.4197 & 0.1834 & 0.1034 & 0.1598 & 0.0702 & 1.0218 \\
		\noalign{\smallskip}
  	       Summation & \scriptsize{\color{green}\CheckmarkBold} & 0.4364 & 0.1908 & 0.1079 & 0.1640 & 0.0722 & 1.0302 \\
            \noalign{\smallskip}   
	       Gated~\cite{arevalo2017gmu} & \scriptsize{\color{green}\CheckmarkBold} & 0.2211 & 0.1011 & 0.0863 & 0.1443 & 0.0629 & 0.9474 \\
		\noalign{\smallskip}        
	       AFF~\cite{dai2021attentional} & \scriptsize{\color{green}\CheckmarkBold} & 0.4602 & 0.2016 & 0.1142 &0.1564 & 0.0686 & 0.9913 \\
		\noalign{\smallskip}
  	       iAFF~\cite{dai2021attentional} & \scriptsize{\color{green}\CheckmarkBold} & 0.4142 & 0.1820 & 0.1130 & 0.1594 & 0.0702 & 1.0112 \\
		\noalign{\smallskip}
  	       BUTD~\cite{anderson2018bottom} & \scriptsize{\color{green}\CheckmarkBold} & 0.1952 & 0.0913 & \textbf{0.0832} & 0.1415 & 0.0610 & 0.9223 \\
            \noalign{\smallskip}
  	        Vision-only & \scriptsize{\color{red}\XSolidBrush} & 0.4857 & 0.3166 & 0.2370 & 0.1363 & 0.0555 & 0.4905\\
            \noalign{\smallskip}
		   V$^2$-SfMLearner & \scriptsize{\color{green}\CheckmarkBold} & \textbf{0.1606} & \textbf{0.0880} & 0.1450 & \textbf{0.1269} & \textbf{0.0521} & \textbf{0.4592} \\
            \bottomrule[1pt]
	\end{tabular}} 
\end{table}

\begin{table}[t]
    \renewcommand\arraystretch{0.4}
    \footnotesize 
	\caption{
	Robustness study with different image corruption methods of our proposed solution against Af-SfMLearner~\cite{AF-SfMLearner}.
        We process the test data with 12 corruption methods, and each image corruption method corresponds to 5 severity levels.
 }
	\centering
	\label{tab:robustness_method}  
        \resizebox{0.48\textwidth}{!}{
	\begin{tabular}{l|cccc|cccc}
        \noalign{\smallskip}\bottomrule[1pt]\noalign{\smallskip}
        \multirow{2}{*}{AF-SfMLearner} & \multicolumn{4}{c|}{w/o Vibration-Vision Fusion} & \multicolumn{4}{c}{w Vibration-Vision Fusion} \\ \noalign{\smallskip}\cline{2-9}\noalign{\smallskip}      
        & \multicolumn{2}{c}{Depth} & \multicolumn{2}{c|}{Ego-motion} & \multicolumn{2}{c}{Depth} & \multicolumn{2}{c}{Ego-motion} \\
        \noalign{\smallskip}\hline\noalign{\smallskip}
        \multicolumn{1}{c|}{Corruption} & AbsDiff $\downarrow$ & Acc $\uparrow$ & ATE $\downarrow$ & ARE $\downarrow$ & AbsDiff $\downarrow$ & Acc $\uparrow$ & ATE $\downarrow$ & ARE $\downarrow$ \\
	\noalign{\smallskip}\hline\noalign{\smallskip}

        Impulse Noise & 0.3185 & 84.55 & 1.0404 & \textbf{0.1775} & \textbf{0.3079} & \textbf{85.99} & \textbf{0.3428} & 0.1786 \\\noalign{\smallskip}
        Speckle Noise & 0.3245 & 83.80 & 1.0520 & 0.1774 & \textbf{0.3136} & \textbf{85.11} & \textbf{0.3327} & \textbf{0.1751} \\\noalign{\smallskip}
        Shot Noise & 0.3271 & 83.60 & 1.0698 & 0.1783 & \textbf{0.3158} & \textbf{84.97} & \textbf{0.3952} & \textbf{0.1772} \\\noalign{\smallskip}
        Gaussian Noise & 0.3261 & 84.03 & 1.0092 & 0.1783 & \textbf{0.3129} & \textbf{85.58} & \textbf{0.4059} & \textbf{0.1776} \\\noalign{\smallskip}
        Glass Blur & 0.3260 & 84.88 & 1.0444 & 0.1758 & \textbf{0.3040} & \textbf{86.73} & \textbf{0.2251} & \textbf{0.1656} \\\noalign{\smallskip}
        Gaussian Blur & 0.3296 & 84.74 & 0.9946 & 0.1752 & \textbf{0.3113} & \textbf{86.23} & \textbf{0.2958} & \textbf{0.1652} \\\noalign{\smallskip}
        Defocus Blur & 0.3303 & 84.70 & 0.9785 & 0.1753 & \textbf{0.3125} & \textbf{86.14} & \textbf{0.3361} & \textbf{0.1652} \\\noalign{\smallskip}
        Saturate & 0.3086 & 85.88 & 0.7477 & 0.1687 & \textbf{0.2948} & \textbf{87.39} & \textbf{0.1736} & \textbf{0.1628} \\\noalign{\smallskip}
        Brightness & 0.3441 & 82.43 & 0.8487 & 0.1712 & \textbf{0.3327} & \textbf{83.37} & \textbf{0.1191} & \textbf{0.1625} \\\noalign{\smallskip}
        Contrast & 0.3297 & 83.59 & 0.5762 & 0.1606 & \textbf{0.3181} & \textbf{84.78} & \textbf{0.1599} & \textbf{0.1565} \\\noalign{\smallskip}
        Spatter & 0.3188 & 84.51 & 0.9264 & 0.1743 & \textbf{0.3156} & \textbf{84.87} & \textbf{0.4773} & \textbf{0.1718} \\\noalign{\smallskip}
        Elastic Transform & 0.3249 & 84.91 & 1.0084 & 0.1776 & \textbf{0.2998} & \textbf{86.99} & \textbf{0.3443} & \textbf{0.1678} \\\noalign{\smallskip}
        \hline\noalign{\smallskip}
        Overall & 0.3257 & 84.30 & 0.9414 & 0.1742 & \textbf{0.3116} & \textbf{85.68} & \textbf{0.3006} & \textbf{0.1688}  \\
        \bottomrule[1pt]\noalign{\smallskip}
	\end{tabular}}
\end{table}

\section{Conclusion \& Discussion}
\label{sec:5}

In this paper, we present V$^2$-SfMLearner, a vision-vibration solution to learn WCE depth and ego-motion. A multimodal learning framework is employed to defend against perturbations from vibrations and collisions in WCE. The proposed Fourier heterogeneous fusion module fuses the output feature spaces from the vision and vibration branches, and finally feeds the multi-task prediction decoders. Key clinical advantages of our multimodal framework include the following: (1) The proposed framework has excellent compatibility with existing vision-only algorithms. (2) Collecting GT in real scenes is difficult and expensive. Our unsupervised solution achieves impressive estimation results without requiring GT data for supervision during training. Meanwhile, a large virtual multimodal dataset is collected, which can help alleviate the problem of complicated and expensive data collection in real scenarios. (3) There have been studies on integrating IMUs into WCE~\cite{abu2015vision, li2022external}. Our solution can achieve superior performance and robustness without any external hardware assistance. Therefore, the proposed framework holds great promise for clinical WCE solutions. In future work, we shall aim to integrate this system with practical clinical applications to meet the actual needs of clinicians. The potential future work and challenges may include: (1) reducing network redundancy and achieving exceptional performance even with limited computing resources; (2) further generalizing our method to real datasets using Sim2Real techniques. 

Besides, the proposed method may extend beyond WCE to various endoscopic procedures, such as laparoscopic surgeries, bronchoscopy, and colonoscopy, which face challenges like insufficient lighting, low-contrast visuals, and motion artifacts. By integrating vibration signals from embedded IMUs or external sensors, the framework enhances depth estimation and motion tracking for more robust performance in these scenarios. Additionally, vibration artifacts, common in robotic-assisted surgeries and flexible endoscopy due to instrument movements or patient biomechanics, are addressed through the framework's vibration network branch and Fourier heterogeneous fusion module, which mitigate vibration noise and improve visual data quality.

\bibliographystyle{IEEEtran}
\bibliography{reference}

\begin{IEEEbiography}
[{\includegraphics[width=1in,height=1.25in,clip,keepaspectratio]{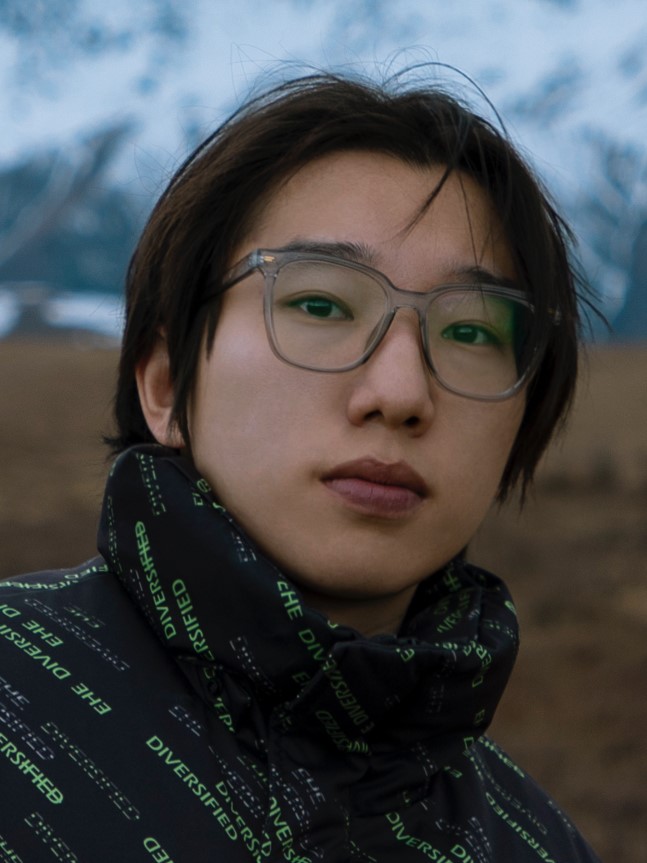}}]{Long Bai} 
received the B.S. degree in Opto-Electronics Information Science and Engineering from the Beijing Institute of Technology (BIT) in 2021. He is currently pursuing the Ph.D. degree with the Department of Electronic Engineering, The Chinese University of Hong Kong (CUHK); and is also a visiting Ph.D. student at the Chair for Computer Aided Medical Procedures \& Augmented Reality (CAMP), Technical University of Munich (TUM), advised by Prof. Nassir Navab. 
He is a recipient of the CUHK Vice-Chancellor's Ph.D. Scholarship Scheme (2021), ICBIR Best Student Paper Award (2023), ICRA Workshop Best Poster Awards (2023 \& 2024), IPCAI Best Paper Award Shortlist (2024), MICCAI Best Paper Runner Up (2024). 
His current research interests include vision-language learning, robotic perception, and surgical data science.
\end{IEEEbiography}

\begin{IEEEbiography}
[{\includegraphics[width=1in,height=1.25in,clip,keepaspectratio]{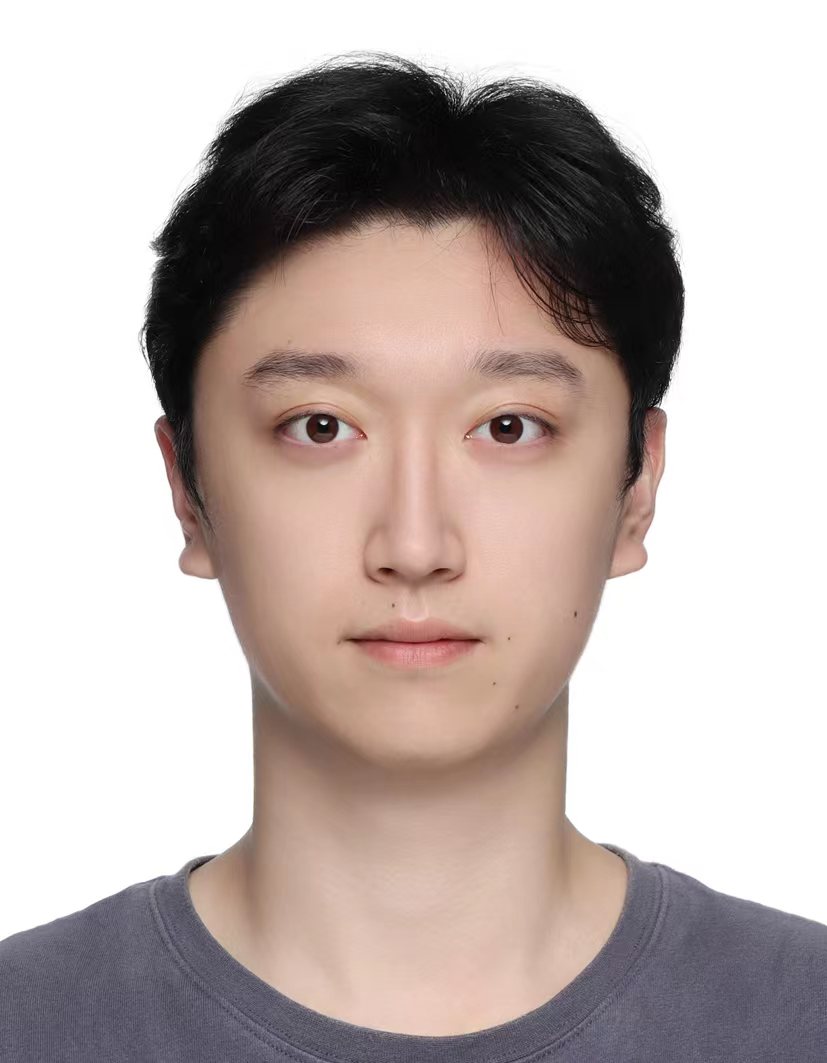}}]{Beilei Cui} received the B.E. degree in Communication Engineering from the University of Electronic Science and Technology of China, Chengdu, China, in 2021, and the M.S. degree in Electronic Engineering from Imperial College London, London, UK, in 2022. He is currently pursuing the Ph.D. degree with the Department of Electronic Engineering, The Chinese University of Hong Kong. 
His current research interests include depth estimation, surgical scene reconstruction, and machine learning in medical robotics.
\end{IEEEbiography}

\begin{IEEEbiography}
[{\includegraphics[width=1in,height=1.25in,clip,keepaspectratio]{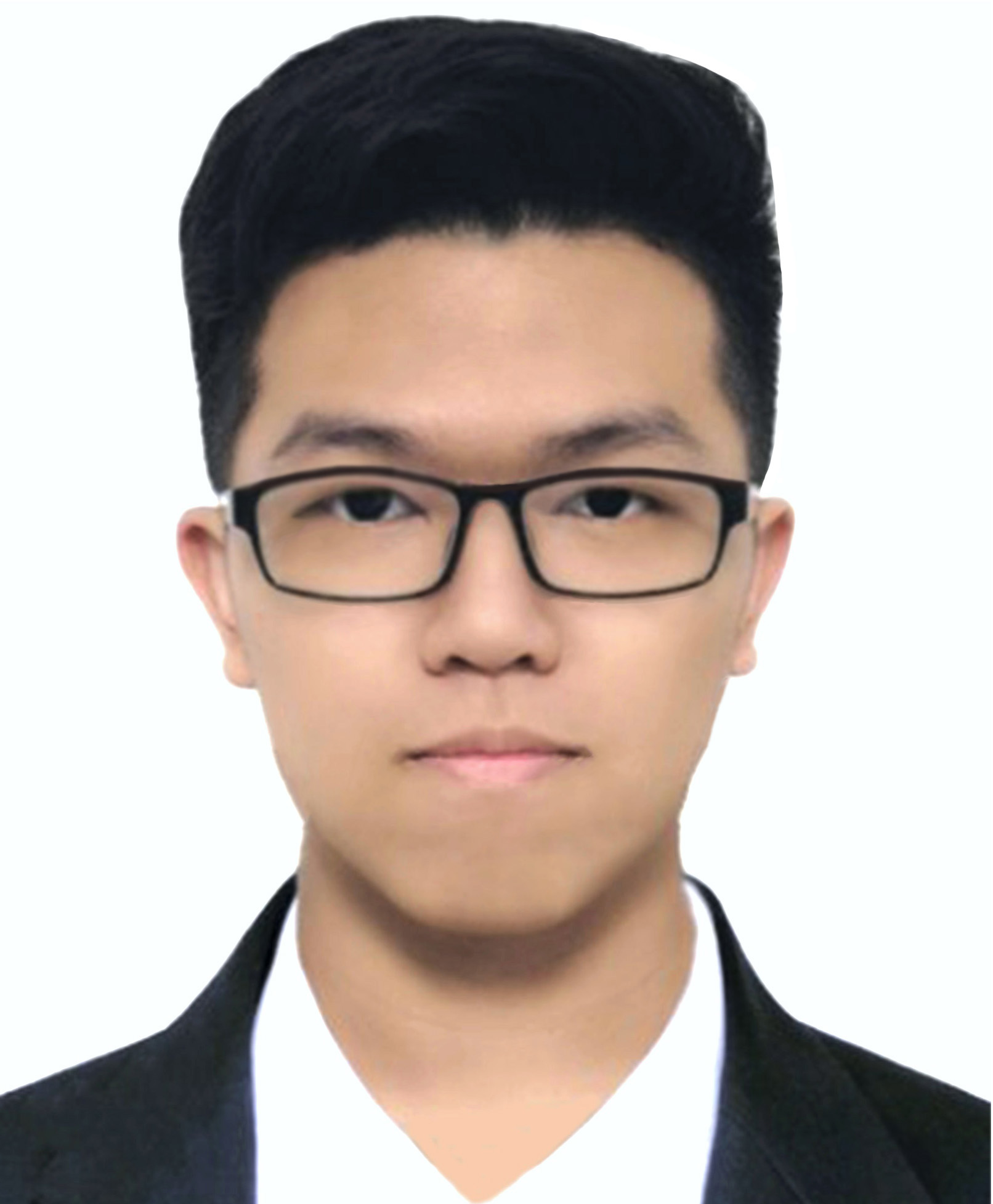}}]{Liangyu Wang} received the B.S. degree in Communication Engineering from Tianjin University, Tianjin, China, in 2021, and the M.S. degree in Electronic Engineering from The Chinese University of Hong Kong, Hong Kong, China, in 2022. 
He is now pursuing his Ph.D. degree at King Abdullah University of Science and Technology, Saudi Arabia.
His research interests include multimodal learning and fair federated learning.
\end{IEEEbiography}

\begin{IEEEbiography}
[{\includegraphics[width=1in,height=1.25in,clip,keepaspectratio]{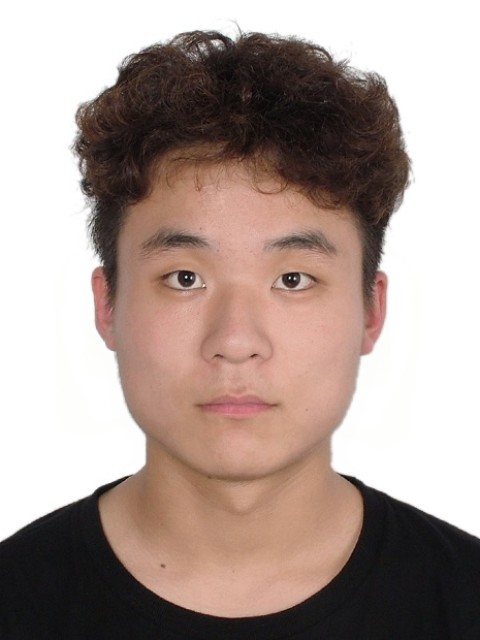}}]{Shilong Yao}
received the B.E. degree in Mechanical Engineering from the Southern University of Science and Technology, Shenzhen, China, in 2021. He is currently pursuing the Ph.D. degree with the Department of Electrical Engineering, City University of Hong Kong (CityU), Hong Kong. 

His current research interests include surgical robotic systems and machine learning in medical robotics.
\end{IEEEbiography}

\begin{IEEEbiography}
[{\includegraphics[width=1in,height=1.25in,clip,keepaspectratio]{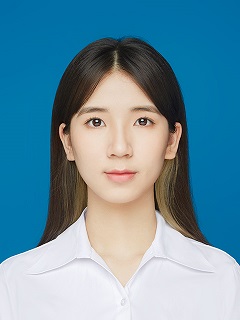}}]{Yanheng Li} received the B.S. degree in Industrial Design from Beijing Institute of Technology (BIT), Beijing, China, in 2022. She is currently pursuing the Ph.D. degree at the School of Creative Media, City University of Hong Kong (CityU), Hong Kong, supported by Hong Kong PhD Fellowship Scheme (HKPFS).
Her current research interests include embodied human-computer interaction, human-robot interaction, and AI for well-being.
\end{IEEEbiography}

\begin{IEEEbiography}
[{\includegraphics[width=1in,height=1.25in,clip,keepaspectratio]{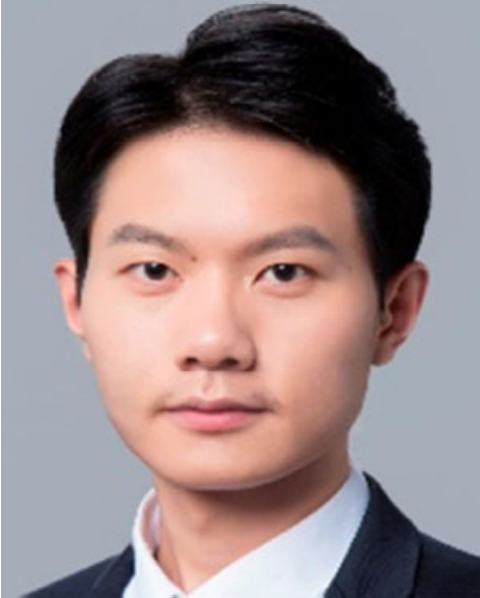}}]{Sishen Yuan}
received the B.S. degree in Mechanical Design Manufacture and Automation from the Harbin Institute of Technology (HIT) at Weihai, Weihai, China, in 2018, and the M.S. degree in Mechatronic Engineering from the Harbin Institute of Technology (HIT) at Shenzhen, Shenzhen, China, in 2021. He is currently pursuing the Ph.D. degree with the Department of Electronic Engineering, The Chinese University of Hong Kong (CUHK), Hong Kong.
His current research interests include magnetic manipulated medical robots covering interventional flexible robotics, capsule endoscopy, and the design and actuation of magnetic-driven origami robots.

\end{IEEEbiography}

\begin{IEEEbiography}
[{\includegraphics[width=1in,height=1.25in,clip,keepaspectratio]{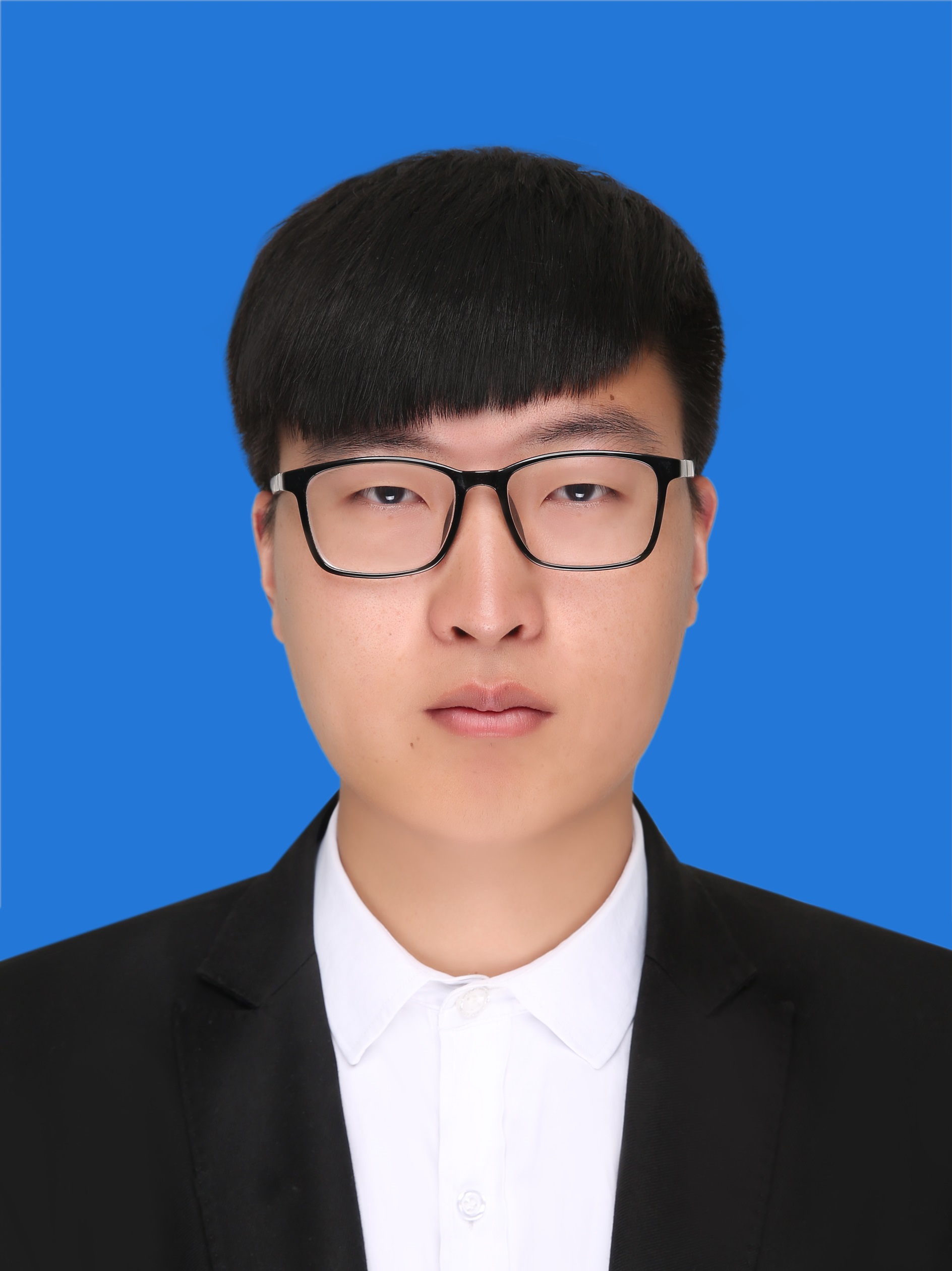}}]{Yanan Wu}
is currently a lecturer at the School of Health Management, China Medical University. He received the B.S. degree in Electronic Engineering from the Huzhou University, Huzhou, China, in 2017, the M.S. degree in Biomedical Engineering from the China Medical University (CMU), Shenyang, China, in 2020, and the Ph.D. degree in Biomedical Engineering from the Northeastern University (NEU), Shenyang, China, in 2024.
His current research interests include machine learning in lung diseases.
\end{IEEEbiography}

\begin{IEEEbiography}
[{\includegraphics[width=1in,height=1.25in,clip,keepaspectratio]{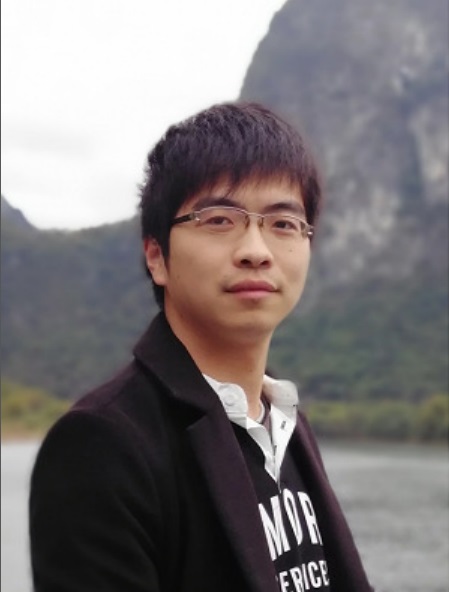}}]{Yang Zhang} received the Ph.D. degree in computer science and technology from the National Key Laboratory for Novel Software Technology, Department of Computer Science and Technology, Nanjing University, Nanjing, China, in 2021. He is currently an Assistant Professor with the School of Mechanical Engineering, Hubei University of Technology, Wuhan, China. His current research interests include machine learning and computer vision.
\end{IEEEbiography}

\begin{IEEEbiography}
[{\includegraphics[width=1in,height=1.25in,clip,keepaspectratio]{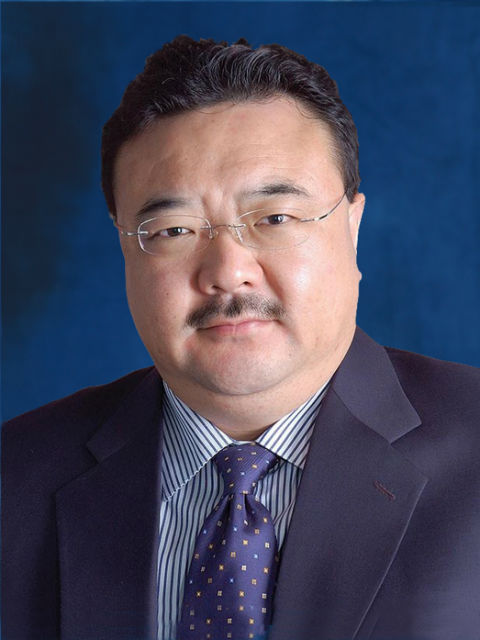}}]{Max Q.-H. Meng}
(Fellow, IEEE) received the Ph.D. degree in electrical and computer engineering from the University of Victoria, Victoria, BC, Canada, in 1992.
He is currently a Chair Professor and the Head of the Department of Electronic and Electrical Engineering with the Southern University of Science and Technology, Shenzhen, China. 
He joined the Chinese University of Hong Kong in 2001 as a Professor and later the Chairman of Department of Electronic Engineering. He was with the Department of Electrical and Computer Engineering, University of Alberta, Edmonton, AB, Canada, where he was the Director of the Advanced Robotics and Teleoperation Lab and held the positions of Assistant Professor in 1994, an Associate Professor in 1998, and a Professor in 2000. 
He has authored or coauthored more than 750 journal and conference papers and book chapters and led more than 60 funded research projects to completion as a Principal Investigator. 
His research interests include robotics, perception, and intelligence.

Prof. Meng is a fellow of the Hong Kong Institution of Engineers and an Academician of the Canadian Academy of Engineering and an Elected Member of the AdCom of IEEE RAS for two terms. He was a recipient of the IEEE Millennium Medal. He is the General Chair or Program Chair of many international conferences, including the General Chair of IROS 2005 and ICRA 2021, respectively. He served as an Associate VP for Conferences of the IEEE Robotics and Automation Society from 2004 to 2007 and the Co-Chair of the Fellow Evaluation Committee. He is the Editor-in-Chief and with the editorial board of a number of international journals, including the Editor-in-Chief of the Elsevier Journal of \emph{Biomimetic Intelligence and Robotics}.
\end{IEEEbiography}

\begin{IEEEbiography}
[{\includegraphics[width=1in,height=1.25in,clip,keepaspectratio]{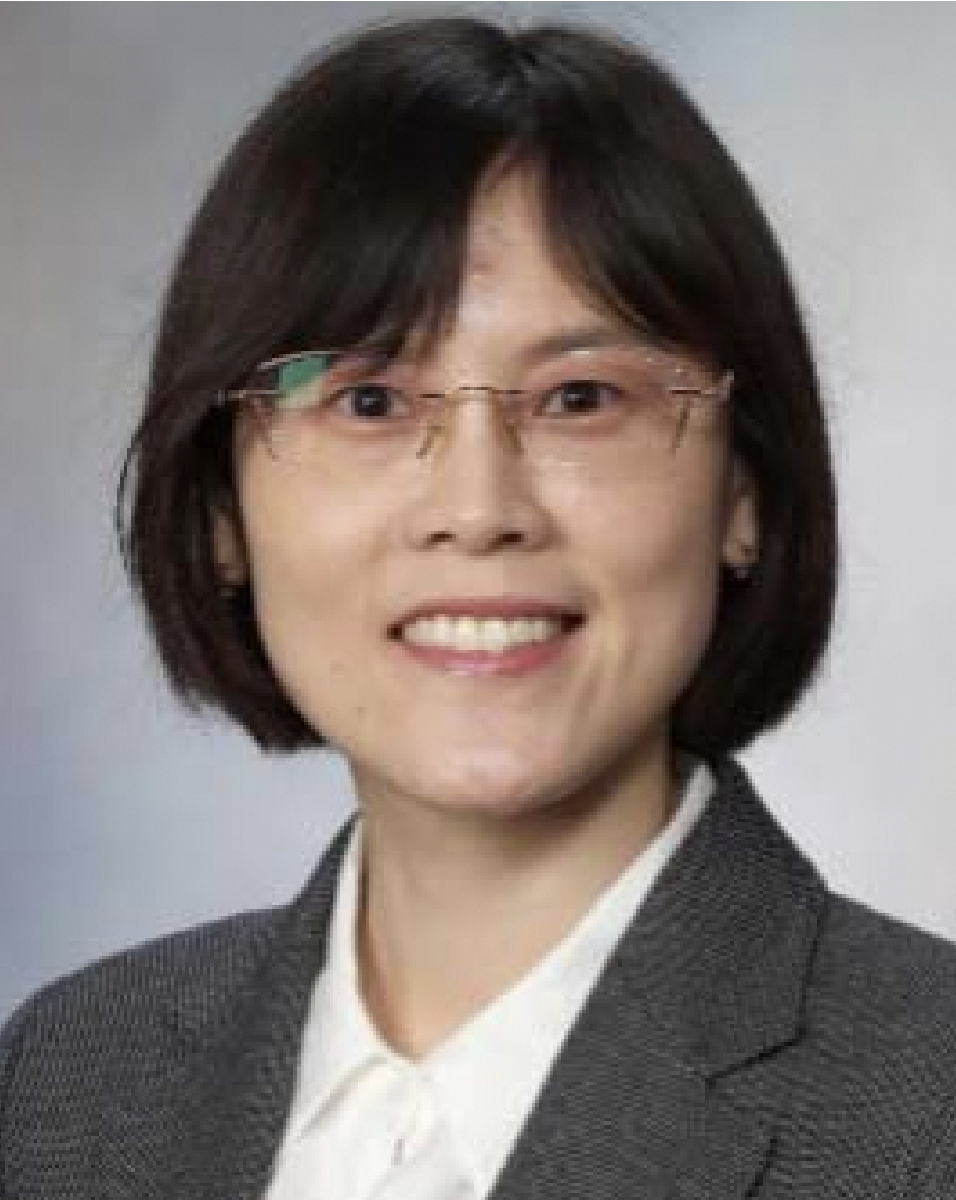}}]{Zhen Li} is an Associate Professor in the Department of Gastroenterology at Qilu Hospital of Shandong University. She worked as a visiting scientist as Mayo Clinic Florida from 2021 to 2022. She is an experienced endoscopist in performing advance endoscopic therapies including EUS, ESD, POEM, and ERCP. She is the mentor of postgraduate students and is the principal investigator of several clinical trials in the field of GI cancer screening. 
Her research areas include endoscopic screening and surveillance of early gastrointestinal neoplasia, microbiota changes of GI diseases, and basic mechanisms of intestinal barrier damage. She has published more than 40 SCI papers in these research areas, and has achieved multiple fundings from the national and provincial scientific grants.
\end{IEEEbiography}

\begin{IEEEbiography}
[{\includegraphics[width=1in,height=1.25in,clip,keepaspectratio]{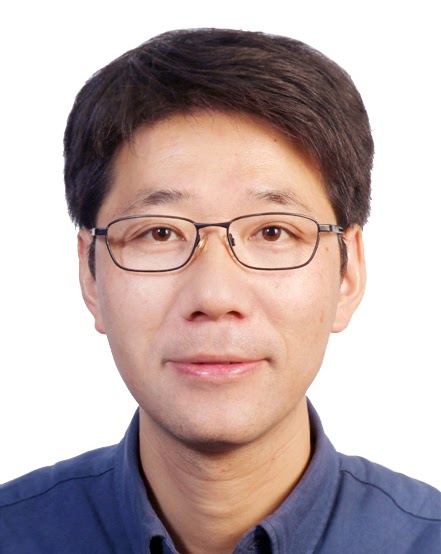}}]{Weiping Ding} (Senior Member, IEEE) received the Ph.D. degree in computer science from the Nanjing University of
Aeronautics and Astronautics, Nanjing, China, in 2013.

He is currently a Full Professor and Dean with the School of Artificial Intelligence and Computer Science, Nantong University, Nantong, China. He was a Postdoctoral Researcher with National Chiao Tung University, Hsinchu, Taiwan from 2014 to 2015, a Visiting Scholar with the National University of Singapore, Singapore, in 2016, and a Visiting Professor with the University of Technology Sydney, Sydney, NSW, Australia, from 2017 to 2018. He has published over 350 articles, including 150 IEEE Transactions papers, and co-authored five books. He holds over 50 approved invention patents, including U.S. and Australian ones. His research focuses on deep neural networks, granular data mining, and multimodal machine learning.

Dr. Ding has recognized as a top 2\% scientist by Stanford University from 2020 to 2024. He serves on editorial boards for numerous IEEE journals and is the leading guest editor of special issues in prestigious publications. He is currently a Co-Editor-in-Chief of three journals, including \textit{Journal of Artificial Intelligence and Systems}, \textit{Journal of Artificial Intelligence Advances}, and \textit{Sustainable Machine Intelligence Journal}.
\end{IEEEbiography}

\begin{IEEEbiography}
[{\includegraphics[width=1in,height=1.25in,clip,keepaspectratio]{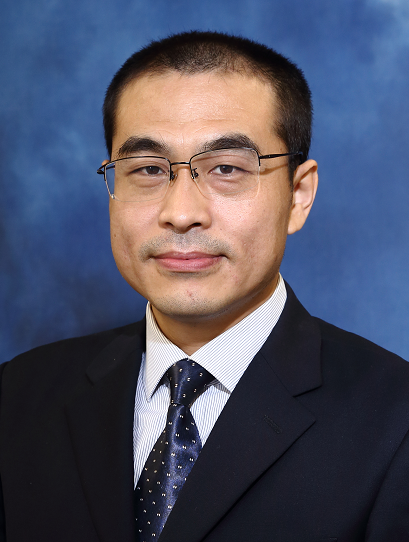}}]{Hongliang Ren}
(Senior Member, IEEE) received the Ph.D. degree in electronic engineering (specialized in biomedical engineering) from The Chinese University of Hong Kong (CUHK) in 2008. He has navigated his academic journey through CUHK, Johns Hopkins University, Children’s Hospital Boston, Harvard Medical School, Children’s National Medical Center, USA, and National University of Singapore (NUS). He is currently a Professor with the Department of Electronic Engineering, CUHK. 
His research interests include biorobotics, intelligent control, medical mechatronics, soft continuum robots, soft sensors, and multisensory learning in medical robotics.

He was the recipient of the IFMBE/IAMBE Early Career Award 2018, Interstellar Early Career Investigator Award 2018, ICBHI Young Investigator Award 2019, and the Health Longevity Catalyst Award 2022 by NAM and RGC. He has served as an Active Organizer and Contributor on the committees of numerous robotics conferences, including a variety of roles in the flagship IEEE International Conference on Robotics and Automation, IEEE/RSJ International Conference on Intelligent Robots and Systems, as well as other domain conferences, such as ROBIO/BIOROB/ICIA. He served as an Associate Editor for \emph{IEEE Transactions on Automation Science and Engineering} and \emph{Medical \& Biological Engineering \& Computing (MBEC)}.
\end{IEEEbiography}

\vfill

\end{document}